\documentclass[11pt]{article}

\usepackage[utf8]{inputenc} 
\usepackage[T1]{fontenc}    
\usepackage{hyperref}       
\usepackage{url}            
\usepackage[numbers,sort&compress]{natbib} 
\usepackage{booktabs}       
\usepackage{amsfonts}       
\usepackage{amsmath}        
\usepackage{nicefrac}       
\usepackage{fancyhdr}       
\usepackage{graphicx}       
\usepackage{algorithm}      
\usepackage{algorithmic}    
\usepackage{geometry}       
\usepackage{titlesec}       

\usepackage{float}          
\usepackage{placeins}
\usepackage[font=small]{caption}
\usepackage{etoolbox} 

\graphicspath{{./}{./figures/}{./ablation_figures/}{./assets/}}     

\fancypagestyle{firstpage}{
    \fancyhf{}
    \fancyhead[L]{%
        \includegraphics[height=1.0cm,keepaspectratio]{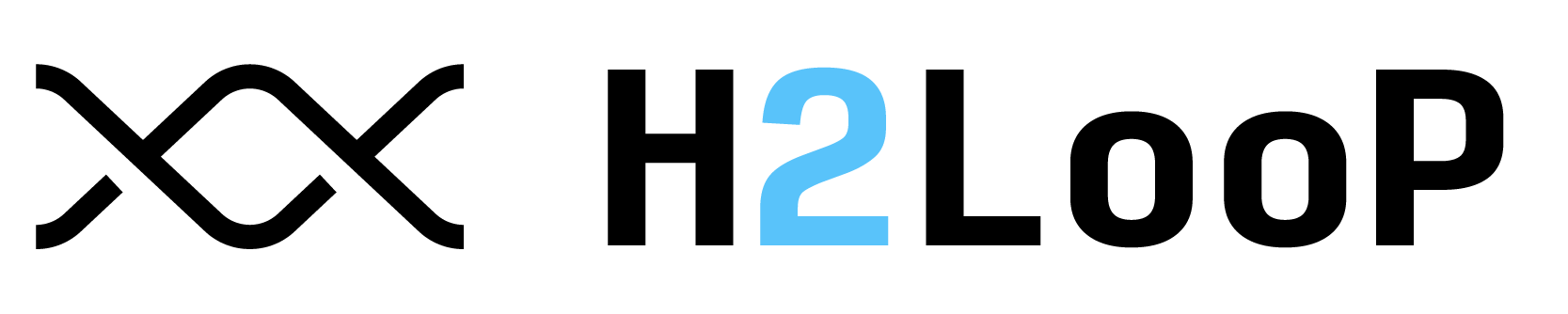}
    }
    
}

\makeatletter
\renewcommand{\maketitle}{
  \begin{center}
    {\LARGE\bfseries \@title \par}
    \vskip 1.5em
    {\large \@author \par}
  \end{center}
  \vskip 2em
}
\makeatother

\titleformat{\section}
  {\Large\bfseries}{\thesection}{1em}{}
\titleformat{\subsection}
  {\large\bfseries}{\thesubsection}{1em}{}
\titleformat{\subsubsection}
  {\normalsize\bfseries}{\thesubsubsection}{1em}{}

\title{H2LooP Spark Preview: Continual Pretraining of Large Language Models for Low-Level Embedded Systems Code}

\author{
\begin{tabular}{c c c c c}
Amit Singh & Vedant Nipane & Pulkit Agrawal & Jatin Kishnani & Sairanjan Mishra  \\
\end{tabular}
\\[0.5em]
\textbf{H2LooP.ai}
}

\begin{document}
\thispagestyle{firstpage}

\maketitle

\begin{abstract}
Large language models (LLMs) have demonstrated remarkable code generation capabilities across general-purpose programming languages, yet they remain fundamentally limited in specialized domains such as low-level embedded systems programming. This domain is characterized by direct hardware register manipulation, vendor-specific SDK patterns, real-time operating system APIs, and hardware abstraction layers that are severely underrepresented in standard pretraining corpora. We present \textbf{H2LooP Spark Preview}, a continual pretraining (CPT) pipeline that adapts OLMo-3-7B---a 7-billion parameter open-weight language model---to the embedded systems domain using BF16 LoRA with Rank-Stabilized scaling on 8$\times$ NVIDIA H100 80GB GPUs. Our training corpus is constructed from 818 repository-datasheet pairs spanning 76.4 GB of raw embedded systems data across 117 manufacturers and 19 component categories, processed using the hierarchical datasheet-to-code mapping methodology introduced in SpecMap (Nipane et al., 2026 ~\cite{nipane2026specmap}). The resulting curated corpus contains approximately 23.5 billion tokens organized across 13 evaluation domains. Through systematic hyperparameter exploration across 4 Weights \& Biases projects encompassing over 1,400 individual runs (30 substantive runs exceeding 2 hours, totaling $\sim$915 wall-clock hours and $\sim$4,240 GPU-hours)---including Bayesian optimization sweeps and grid searches over LoRA rank, target modules, learning rates, and data mixture ratios---we identify that high-rank LoRA ($r$=512) with conservative learning rates and full-module targeting provides optimal domain adaptation. After training on 5.2 billion tokens ($\sim$36\% of one epoch), Spark Preview achieves a \textbf{70.4\% reduction in in-domain perplexity} (4.06 $\to$ 1.20) and a \textbf{66.1\% reduction on held-out repositories} (3.92 $\to$ 1.33) not seen during training. On generative code completion benchmarks across 13 embedded domains, our 7B-parameter model surpasses Claude Opus 4.6 and Qwen3-Coder-30B on \textbf{8 of 13 categories} in token accuracy, demonstrating that targeted continual pretraining enables small open-weight models to rival frontier systems on specialized technical tasks. We release a checkpoint of the production training run as open-source at \textbf{\href{https://huggingface.co/h2loop-ai/spark-cpt-base-ckpt}{spark-cpt-base-ckpt}}.

\end{abstract}

\vskip 1em

\section{Introduction}

The embedded systems software ecosystem represents one of the most demanding domains for automated code understanding and generation. Unlike application-level software where patterns are well-represented in web-scale training corpora, embedded code operates at the hardware-software boundary---a space defined by vendor-specific register maps, interrupt handling conventions, peripheral initialization sequences, and real-time constraints. A typical embedded project may span thousands of files implementing communication protocols from hardware datasheets, with the corresponding specification documents containing hundreds of sections describing interfaces, data structures, and behavioral requirements (Nipane et al., 2026)~\cite{nipane2026specmap}.

The challenges of embedded systems code are multifaceted. First, \textbf{hardware specificity}: each microcontroller family (STM32, AURIX, S32K, MSP430, and others) defines its own register layout, peripheral configuration API, and startup sequence. A valid register write for an STM32F4 CAN peripheral is nonsensical for an NXP FlexCAN module, despite both implementing the same communication protocol. Second, \textbf{code artifact diversity}: embedded implementations involve not just functions, but macros, structs, constants, enums, typedefs, preprocessor definitions, device tree bindings, linker scripts, and hardware abstraction layers---artifact types that general code models rarely encounter in training. Third, \textbf{scale and fragmentation}: the embedded ecosystem spans hundreds of manufacturers, thousands of component variants, and millions of lines of vendor SDK code, much of which exists only in proprietary documentation and reference implementations.

General-purpose LLMs trained on broad web corpora---including frontier models like Claude Opus 4.6 and large coding models like Qwen3-Coder-30B (Yang et al., 2025)~\cite{yang2025qwen3}---have limited exposure to these specialized patterns. When asked to generate embedded code, they frequently hallucinate register names, produce incorrect peripheral initialization sequences, confuse vendor-specific API conventions, and violate hardware-specific timing constraints. These failures stem not from a lack of reasoning capability, but from insufficient exposure to the domain during pretraining.

\textbf{Continual pretraining} (CPT) offers a principled solution: by continuing the language modeling objective on domain-specific data while using parameter-efficient fine-tuning (PEFT) methods, we can inject domain knowledge into a pretrained model without catastrophic forgetting of its broad capabilities. This approach has proven effective in other domains, but its application to embedded systems remains largely unexplored.

In this work, we present \textbf{H2LooP Spark Preview}, a comprehensive CPT pipeline for low-level embedded systems code. Our contributions are:

\begin{enumerate}
\item \textbf{A large-scale embedded systems training corpus} constructed from repository-datasheet pairs ($>$100B raw tokens) spanning 117 manufacturers and 61 component classes, processed using the SpecMap hierarchical mapping methodology (Nipane et al., 2026)~\cite{nipane2026specmap} for datasheet-to-code traceability. A novel approach which enabled us to defragment related artifacts scattered across web-crawl dumps and open internet.

\item \textbf{Bayesian hyperparameter exploration} across 1,400+ runs ($\sim$4,240 GPU-hours), establishing that high-rank LoRA ($r$=512) with RSLoRA scaling, conservative learning rates, and full-module targeting (attention + MLP + embeddings) is optimal for domain CPT, while demonstrating that domain-only data outperforms mixed corpora across all data mixtures.

\item \textbf{Evaluation framework} spanning perplexity, teacher-forced completion accuracy, and free-form generative benchmarks on both in-domain data and in-production held-out GitHub repositories, evaluated across 13 specialized embedded domains.

\item \textbf{Demonstration that a 7B CPT model rivals frontier models}: Spark Preview achieves the highest generative token accuracy on 8 of 13 embedded categories, outperforming Claude Opus 4.6 (a leading frontier coding model) and Qwen3-Coder-30B (Yang et al., 2025)~\cite{yang2025qwen3} despite being 4$\times$ smaller.

\item \textbf{Open-source release} of a production checkpoint at \textbf{\href{https://huggingface.co/h2loop-ai/spark-cpt-base-ckpt}{spark-cpt-base-ckpt}} to enable community research on embedded systems LLMs under a Research Only License.
\end{enumerate}

\section{Related Work}

\subsection{Continual Pretraining for Domain Adaptation}

Continual pretraining extends a model's knowledge by continuing the language modeling objective on domain-specific corpora. This approach has been successfully applied across multiple domains. \textbf{CodeLlama} (Roziere et al., 2023)~\cite{roziere2023codellama} applied CPT to Llama 2 with 500B additional tokens of code, producing a family of code-specialized models that substantially outperformed the general-purpose base. \textbf{BioMedLM} (Bolton et al., 2024)~\cite{bolton2024biomedlm} demonstrated that CPT on biomedical literature produces models competitive with much larger general-purpose systems on medical question answering. ChipNeMo (Liu et al., 2024)~\cite{liu2024chipnemo} represents the closest prior work to ours, applying CPT to LLaMA2 models for chip design tasks including EDA script generation and bug summarization, though their focus is on digital ASIC design rather than embedded systems firmware.

Our work differs from these in several key aspects: (1) the embedded systems domain requires understanding of hardware-software co-design patterns that span datasheets, firmware, and hardware abstraction layers; (2) our training data is constructed through a structured datasheet-to-code mapping process rather than simple web crawling; and (3) we evaluate on a broader range of domain-specific categories (13 distinct embedded specializations).

\subsection{Parameter-Efficient Fine-Tuning}

\textbf{Low-Rank Adaptation (LoRA)} (Hu et al., 2022)~\cite{hu2022lora} introduces trainable low-rank decomposition matrices into frozen pretrained weights, enabling fine-tuning with dramatically reduced memory requirements. \textbf{QLoRA} (Dettmers et al., 2023)~\cite{dettmers2023qlora} extends this with 4-bit quantization of the base model, further reducing memory usage. \textbf{RSLoRA} (Kalajdzievski, 2023)~\cite{kalajdzievski2023rslora} addresses scaling issues in high-rank LoRA by normalizing the adapter contribution by $1/\sqrt{r}$, preventing the effective learning rate from vanishing with increasing rank. The distinction is captured by the effective learning rate:

\[
\eta_{\text{eff}} = \eta \cdot \frac{\alpha}{r} \quad \text{(standard LoRA)}, \qquad \eta_{\text{eff}} = \eta \cdot \frac{\alpha}{\sqrt{r}} \quad \text{(RSLoRA)}
\]

Under standard LoRA, doubling the rank halves the effective learning rate, requiring compensatory increases in $\alpha$ or $\eta$. RSLoRA decouples rank from effective learning rate scaling, enabling stable training at high ranks without retuning.

The choice of LoRA configuration---particularly rank, target modules, and alpha scaling---significantly impacts CPT outcomes but has received limited systematic study in the context of domain adaptation. Prior work (Biderman et al., 2024)~\cite{biderman2024lora} has shown that LoRA rank requirements vary substantially across tasks and domains. Our work provides the first comprehensive sweep study for embedded systems CPT, testing ranks from 8 to 2048 across multiple module targeting strategies. Through our prior training runs, we have shown how with well-tuned hyperparameters, LoRA training matches (and sometimes exceeds) full training (Singh et al., 2025)~\cite{singh2025convergence}. This work builds from our prior results and also exhibits significantly lower catastrophic forgetting, shown in general benchmarks.

\subsection{Code Language Models}

The landscape of code-specialized LLMs has evolved rapidly. Codex (Chen et al., 2021)~\cite{chen2021codex} and StarCoder (Li et al., 2023)~\cite{li2023starcoder} demonstrated that training on large code corpora produces strong code generation capabilities. CodeBERT (Feng et al., 2020)~\cite{feng2020codebert} and GraphCodeBERT (Guo et al., 2021)~\cite{guo2021graphcodebert} established foundations for code understanding with pre-trained representations. More recently, DeepSeek-Coder Series (Guo et al., 2024)~\cite{guo2024deepseek}, Qwen-Coder Series (Hui et al., 2024)~\cite{hui2024qwen}, and Qwen3 (Yang et al., 2025)~\cite{yang2025qwen3} have pushed the frontier of open-weight code models.

However, these models are trained predominantly on general-purpose languages (Python, JavaScript, Java) with comparatively little representation of systems-level C/C++ code, assembly, device tree source, linker scripts, and hardware-specific header files. As Wang et al. (2024)~\cite{wang2024traceability} demonstrated for traceability link recovery, purely text-based methods have reached their limits for understanding code-specification relationships---LLM-based approaches significantly outperform traditional textual similarity methods. Our work builds on this insight by creating a model specifically adapted to the embedded systems domain.

\subsection{Datasheet-to-Code Mapping for Training Data}

The construction of high-quality domain-specific training data is a critical challenge for CPT. Our work \textbf{SpecMap} (Nipane et al., 2026)~\cite{nipane2026specmap} introduced a hierarchical LLM agent for datasheet-to-code traceability link recovery in embedded systems. This hierarchical approach achieved a practical method for constructing structured training pairs from embedded systems repositories and their associated datasheets. Our training corpus leverages the data pools constructed through this methodology.

\subsection{Embedded Systems AI}

The intersection of AI and embedded systems has received growing attention. Prior work has focused on deploying ML models \emph{on} embedded devices, but the complementary problem of using AI to \emph{understand and generate} embedded code remains underexplored. Industrial embedded systems face unique challenges: safety-critical constraints (ISO 26262, IEC 61508), real-time requirements, and the need for hardware-specific optimization. These constraints make embedded code generation particularly demanding and particularly valuable for automation.

\section{Training Data}
\label{sec:training_data}

\subsection{Data Sources and the Model Training Pool}

Our training corpus is constructed from the \textbf{Model Training Pool}, a curated collection of repository-datasheet pairs assembled through the SpecMap pipeline (Nipane et al., 2026)~\cite{nipane2026specmap}. Each pair links a GitHub repository containing embedded systems code to its corresponding hardware datasheet or reference manual, establishing the structured relationship between specification and implementation.

The Model Training Pool spans:

\begin{itemize}
\item \textbf{Unique repository-datasheet pairs} covering the breadth of the embedded ecosystem
\item \textbf{117 manufacturers} including all major semiconductor vendors
\item \textbf{19 component categories} ranging from sensors and actuators to networking and power management
\item \textbf{61 component classes}
\item \textbf{76.4 GB} of raw repository data
\end{itemize}

\begin{figure}[h]
\centering
\includegraphics[width=\textwidth]{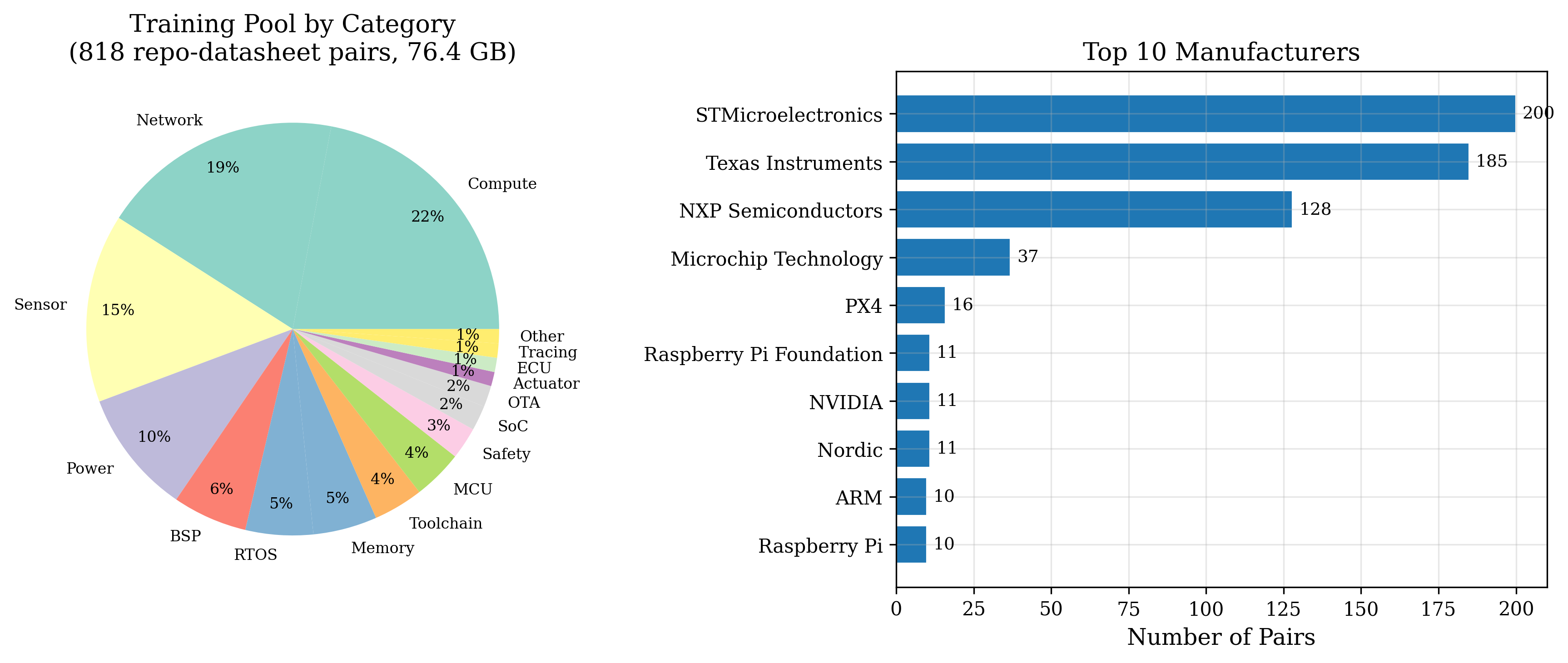}
\caption{Dataset Composition --- Left: Training pool category distribution across 818 pairs. Right: Top 10 manufacturers by pair count.}
\label{fig:dataset_composition}
\end{figure}

The manufacturer distribution reflects the structure of the embedded industry, with \textbf{STMicroelectronics}, \textbf{Texas Instruments}, \textbf{NXP Semiconductors}, and \textbf{Microchip Technology} dominating the corpus. See Figure~\ref{fig:dataset_composition} for the distribution.

\subsection{Data Processing Pipeline}

The raw repository data undergoes a multi-stage processing pipeline to produce a training-ready corpus.

\textbf{Stage 1: Hierarchical Extraction via SpecMap.} Raw mapping files are generated through the SpecMap pipeline (Nipane et al., 2026)~\cite{nipane2026specmap}, which performs hierarchical specification-to-code traceability link recovery. For each repository-datasheet pair, the system executes four sequential phases:

\begin{enumerate}
\item \textbf{Folder Discovery} ($M_1$): Maps datasheet sections to relevant repository folders using LLM-based semantic analysis with adaptive chunking.
\item \textbf{File Discovery} ($M_2$): Identifies specific implementation files within relevant folders through on-demand structure generation and LLM-based relevance scoring.
\item \textbf{Code Symbol Discovery} ($M_3$): Extracts specific code symbols (functions, macros, structs, constants, enums, typedefs) from relevant files using Universal Ctags integration for reliable C/C++ parsing.
\item \textbf{Validation \& Gap Analysis} ($M_4$): Validates the complete mapping sequentially to maintain context across sections, determining implementation status.
\end{enumerate}

This hierarchical decomposition produces structured mapping files that preserve the relationship between datasheet specifications and their code implementations across all levels of the software architecture.

\textbf{Stage 2: Section-Aware Chunking.} The mapping output is processed with 16 parallel workers. Content is split at section boundaries (82-character equals-sign delimiters), with special handling for repository code sections that preserve file boundaries via \texttt{// File:} markers. Large blocks undergo intelligent splitting at code-aware boundaries:

\begin{itemize}
\item \textbf{File boundaries} (preferred): Split at \texttt{// File:} markers
\item \textbf{Function boundaries}: Split at function definitions and closing braces
\item \textbf{Statement boundaries} (fallback): Split at semicolons and double newlines
\end{itemize}

This hierarchy ensures semantic coherence within each training sample. Samples are constrained to a maximum of 7,500 characters ($\sim$2,048 tokens at 3.5 characters per token), with a minimum of 50 characters.

\textbf{Stage 3: Quality Filtering.} Aggressive garbage cleaning removes:

\begin{itemize}
\item Separator lines (repeated dashes, equals, stars, underscores $\geq$ 10 characters)
\item Excessive repetition (any character repeated $\geq$ 10 times, reduced to 3)
\item ASCII art and box-drawing characters
\item Empty comment blocks and excessive Doxygen decoration
\item Carriage returns and tab normalization (tabs $\to$ 4 spaces)
\end{itemize}

Samples exceeding 70\% garbage content after cleaning are rejected. Remaining samples must satisfy at least one of: (a) presence of code indicators (\texttt{\#include}, \texttt{\#define}, \texttt{void}, \texttt{int}, \texttt{struct}, \texttt{typedef}, or control flow keywords), or (b) containing at least 20 words of natural language text.

\textbf{Stage 4: Mixture Assembly.} Cleaned samples are assembled with strict order preservation (no shuffling) to enable reproducible curriculum ordering. Each sample is truncated to a maximum of 2,048 tokens (at 4 characters per token) with boundary-aware splitting at newlines, periods, or semicolons. An end-of-text token (\texttt{<|endoftext|>}) is appended to each sample.

\subsection{Corpus Statistics}

\begin{table}[H]
\centering
\caption{Processed Training Corpus Statistics}
\label{tab:corpus_stats}
\begin{tabular}{lr}
\toprule
Property & Value \\
\midrule
Raw data size & 76.4 GB (818 repo-datasheet pairs) \\
Processed training data & $\sim$20 GB (289 Parquet files) \\
Estimated total samples & $\sim$11.5M \\
Estimated total tokens & $\sim$23.5B \\
Maximum sequence length & 2,048 tokens \\
Average sample length & $\sim$1,450 tokens \\
Packing efficiency & $\sim$95\% fill rate \\
\bottomrule
\end{tabular}
\end{table}

\subsection{Evaluation Data}

We construct two complementary evaluation sets to measure both in-domain performance and generalization to unseen code:

\textbf{In-domain evaluation} (2,270 samples, 13 MB): Stratified samples drawn from the last two Parquet files of each processed category, ensuring temporal separation from training data. Evaluated across all 13 categories.

\textbf{Held-out evaluation} (1.36M samples, 9.5 GB): Code from 9 public GitHub repositories \textbf{never seen during training}, covering the same 13 categories:

\begin{table}[H]
\centering
\caption{Held-Out Evaluation Repositories}
\label{tab:heldout_repos}
\scriptsize
\setlength{\tabcolsep}{3pt}
\begin{tabular}{@{}p{2.8cm}@{\hspace{0.2cm}}p{3.8cm}@{\hspace{0.2cm}}p{5.5cm}@{}}
\toprule
Repository & URL & Categories \\
\midrule
torvalds/linux & github.com/torvalds/\newline linux & linux\_kernel, amd\_gpu\_registers, device\_tree, wireless\_ble\_wifi, usb\_stack \\[1ex]
Infineon/AURIX\_\newline code\_examples & github.com/Infineon/\newline AURIX\_code\_examples & infineon\_aurix \\[1ex]
STMicroelectronics/\newline STM32CubeF4 & github.com/\newline STMicroelectronics/\newline STM32CubeF4 & stm32\_hal \\[1ex]
nxp-mcuxpresso/\newline mcux-sdk & github.com/nxp-\newline mcuxpresso/mcux-sdk & nxp\_imx \\[1ex]
ARM-software/\newline CMSIS\_5 & github.com/ARM-\newline software/CMSIS\_5 & arm\_cortex\_asm \\[1ex]
zephyrproject-rtos/\newline zephyr & github.com/\newline zephyrproject-rtos/\newline zephyr & zephyr\_rtos \\[1ex]
wolfSSL/wolfssl & github.com/wolfSSL/\newline wolfssl & crypto \\[1ex]
Mbed-TLS/mbedtls & github.com/Mbed-\newline TLS/mbedtls & crypto \\[1ex]
hathach/tinyusb & github.com/hathach/\newline tinyusb & usb\_stack \\
\bottomrule
\end{tabular}
\end{table}

Source file types include \texttt{.c}, \texttt{.h}, \texttt{.S}, \texttt{.s}, \texttt{.dts}, \texttt{.dtsi}, and \texttt{.asm}. The \texttt{register\_defines} and \texttt{general} categories are populated via regex matches across all repositories.

\section{Model and Training Methodology}

\subsection{Base Model}

We select \textbf{OLMo-3-1025-7B} (Team OLMo, 2025)~\cite{teamolmo2025olmo3} as our base model. OLMo 3 is a 7-billion parameter autoregressive language model pretrained on a diverse mixture of web text, code, and academic content, with construction targeting long-context reasoning, function calling, coding, instruction following, and knowledge recall. We choose OLMo 3 for three reasons: (1) fully open weights, training data, and training code enable complete reproducibility; (2) the 7B parameter scale provides a practical balance between capability and computational cost; and (3) the model demonstrates competitive general-purpose performance as a foundation for domain adaptation.

\subsection{Parameter-Efficient Fine-Tuning Configuration}

We apply LoRA adapters with Rank-Stabilized scaling (RSLoRA; Kalajdzievski, 2023)~\cite{kalajdzievski2023rslora} to all major parameter groups. In standard LoRA (Hu et al., 2022)~\cite{hu2022lora}, the pretrained weight matrix $W_0 \in \mathbb{R}^{d \times k}$ is augmented with a low-rank update:

\[
W' = W_0 + \frac{\alpha}{r} B A
\]

where $B \in \mathbb{R}^{d \times r}$, $A \in \mathbb{R}^{r \times k}$, and $\alpha$ is a scaling hyperparameter. RSLoRA (Kalajdzievski, 2023)~\cite{kalajdzievski2023rslora} replaces the $1/r$ normalization with $1/\sqrt{r}$ to prevent the effective learning rate from diminishing with increasing rank:

\[
W' = W_0 + \frac{\alpha}{\sqrt{r}} B A
\]

This modification is critical for our high-rank ($r$=512) configuration, as standard LoRA scaling would attenuate the adapter contribution by a factor of $\sqrt{512} \approx 22.6\times$ relative to RSLoRA.

The total number of trainable parameters introduced by LoRA is:

\[
|\theta_{\text{LoRA}}| = r \sum_{m \in \mathcal{M}} \left(d_m^{\text{in}} + d_m^{\text{out}}\right)
\]

where $\mathcal{M}$ is the set of targeted modules and $d_m^{\text{in}}$, $d_m^{\text{out}}$ are the input and output dimensions of module $m$ (each adapter $m$ introduces matrices $A_m \in \mathbb{R}^{r \times d_m^{\text{in}}}$ and $B_m \in \mathbb{R}^{d_m^{\text{out}} \times r}$). For our configuration with $r$=512 targeting all 9 module types across the OLMo-3-7B architecture, this yields approximately 839M trainable parameters ($\sim$12\% of the 7B base model).

We apply adapters to the following module groups:

\begin{itemize}
\item \textbf{Attention layers:} q\_proj, k\_proj, v\_proj, o\_proj
\item \textbf{MLP layers:} gate\_proj, up\_proj, down\_proj
\item \textbf{Embedding layers:} embed\_tokens, lm\_head
\end{itemize}

This ``full-module'' targeting represents the broadest possible LoRA application. Our hyperparameter sweeps (Section 5) demonstrate that this configuration consistently outperforms selective targeting strategies.

\begin{table}[H]
\centering
\caption{LoRA Configuration}
\label{tab:lora_config}
\begin{tabular}{llp{7cm}}
\toprule
Parameter & Value & Rationale \\
\midrule
LoRA rank ($r$) & 512 & Optimal rank-performance tradeoff (Section 5.2) \\
LoRA alpha ($\alpha$) & 1,024 & Standard 2$\times$ rank scaling \\
LoRA dropout & 0.05 & Mild regularization \\
RSLoRA & Enabled & Stable scaling for high-rank adapters \\
Target modules & All (attn + MLP + embed) & Full targeting optimal (Section 5.2) \\
\bottomrule
\end{tabular}
\end{table}

\subsection{Training Configuration}

We optimize the standard causal language modeling objective. Given a sequence of tokens $x = (x_1, x_2, \ldots, x_T)$, the training loss is the average negative log-likelihood:

\[
\mathcal{L}_{\text{CPT}} = -\frac{1}{T}\sum_{t=1}^{T} \log P_\theta(x_t \mid x_{<t})
\]

where $\theta$ comprises the frozen base model parameters plus the trainable LoRA adapter weights. Only the adapter parameters receive gradient updates.

The effective number of tokens processed per optimizer step is:

\[
T_{\text{step}} = B_{\text{device}} \times G_{\text{accum}} \times N_{\text{GPU}} \times L_{\text{seq}} = 4 \times 8 \times 8 \times 2048 = 524{,}288
\]

\begin{table}[H]
\centering
\caption{Training Hyperparameters}
\label{tab:training_config}
\begin{tabular}{lr}
\toprule
Parameter & Value \\
\midrule
Per-device batch size & 4 \\
Gradient accumulation steps & 8 \\
Number of GPUs & 8 $\times$ H100 80GB SXM \\
\textbf{Effective batch size} & \textbf{256} \\
Max sequence length & 2,048 tokens (with packing) \\
Tokens per optimizer step & 524,288 \\
Optimizer & AdamW (dual parameter groups) \\
Main learning rate & 1.5$\times 10^{-5}$ \\
Embedding/LM-head LR & 7.5$\times 10^{-6}$ (0.5$\times$ main) \\
LR scheduler & Cosine with 10\% warmup \\
Max gradient norm & 5.0 \\
Precision & BF16 (full, not quantized) \\
Target epochs & 1 (single pass over $\sim$23.5B tokens) \\
Seed & 3407 \\
\bottomrule
\end{tabular}
\end{table}

The dual optimizer with reduced embedding learning rate (0.5$\times$) was adopted based on preliminary experiments showing that embedding layers benefit from more conservative updates, consistent with findings in ChipNeMo (Liu et al., 2024)~\cite{liu2024chipnemo}.

\subsection{Infrastructure and Hardware}

Training is performed on a single node equipped with 8$\times$ NVIDIA H100 80GB SXM GPUs interconnected via NVLink/NVSwitch, providing 900 GB/s bidirectional bandwidth per GPU pair. The system is backed by 192 CPU cores used for parallel data preprocessing and tokenization. All training data resides on local NVMe storage to eliminate I/O bottlenecks.

\subsection{Distributed Training Architecture}

We employ PyTorch's \textbf{Distributed Data Parallel (DDP)} via HuggingFace Accelerate, with careful attention to initialization ordering and communication optimization.

\textbf{Model loading before process group initialization.} A critical engineering constraint arises from the interaction between model loading and NCCL initialization. Loading large pretrained models triggers internal CUDA operations that can desync across ranks if the NCCL process group is already active. We therefore load the base model on each rank's assigned GPU \emph{before} initializing the Accelerator:

\begin{verbatim}
# Load model BEFORE distributed init to avoid NCCL desync
model = AutoModelForCausalLM.from_pretrained(
    "allenai/OLMo-3-1025-7B",
    device_map={'': local_rank},
    torch_dtype=torch.bfloat16,
    attn_implementation="flash_attention_2",
)
# ... apply LoRA ...
# THEN initialize distributed
accelerator = Accelerator(kwargs_handlers=[
    InitProcessGroupKwargs(timeout=timedelta(hours=100000))
])
\end{verbatim}

This ordering ensures all ranks complete model loading independently before any collective communication occurs. The extended NCCL timeout (100,000 hours) prevents spurious timeout failures during long preprocessing phases.

\textbf{Non-reentrant gradient checkpointing.} We enable gradient checkpointing with \texttt{use\_reentrant=False}, which is required for compatibility with DDP's static graph optimization. Reentrant checkpointing creates variable backward graph topologies across steps, breaking DDP's assumption of consistent gradient bucket ordering.

\textbf{DDP bucket configuration.} DDP gradient synchronization uses 150 MB all-reduce buckets 

(\texttt{ddp\_bucket\_cap\_mb=150}), substantially larger than the PyTorch default of 25 MB. Larger buckets amortize the per-call NCCL overhead by reducing the total number of all-reduce operations from $\sim$200+ (one per parameter group at default bucket size) to approximately 20--30 fused bucket operations per backward pass.

\subsection{H100-Specific Optimizations}

We configure several hardware-specific optimizations to maximize throughput on H100 GPUs:

\textbf{TF32 tensor core arithmetic.} H100 tensor cores support TF32 (19-bit) precision for single-precision matrix multiplications, providing up to 2$\times$ throughput over FP32 with negligible accuracy impact for training workloads. We enable this via:

\begin{verbatim}
torch.backends.cuda.matmul.allow_tf32 = True
torch.backends.cudnn.allow_tf32 = True
torch.backends.cudnn.benchmark = True
torch.set_float32_matmul_precision('high')
\end{verbatim}

Combined with BF16 mixed precision for the LoRA adapters, this configuration achieves near-peak utilization of the H100's 989 TFLOPS BF16 tensor throughput.

\textbf{Flash Attention 2.} We use Flash Attention 2 (Dao, 2023)~\cite{dao2023flashattention2} for all self-attention computation, reducing attention memory from $O(n^2)$ to $O(n)$ and improving wall-clock speed through kernel fusion. This is specified at model load time via \texttt{attn\_implementation="flash\_attention\_2"}.

\textbf{NCCL transport optimizations.} We configure NCCL to exploit the full H100 interconnect topology:

\begin{itemize}
\item \texttt{NCCL\_P2P\_LEVEL=NVL}: NVLink peer-to-peer transfers between GPU pairs
\item \texttt{NCCL\_NET\_GDR\_LEVEL=5}: GPU Direct RDMA for network-attached communication
\item \texttt{NCCL\_IB\_DISABLE=0}: InfiniBand enabled for inter-node communication readiness
\end{itemize}

\textbf{CUDA memory management.} We configure PyTorch's caching allocator with \texttt{expandable\_segments} and a maximum split size of 256 MB (\texttt{max\_split\_size\_mb:256}), reducing memory fragmentation during the dynamic allocation patterns caused by variable-length sequence packing.

\subsection{Dual Optimizer with Embedding Learning Rate Scaling}

We construct a custom AdamW optimizer with two parameter groups that apply different learning rates to embedding and non-embedding parameters:

\begin{table}[h]
\centering
\caption{Optimizer Parameter Groups}
\label{tab:optimizer_groups}
\begin{tabular}{lll}
\toprule
Parameter Group & Learning Rate & Rationale \\
\midrule
Attention + MLP LoRA adapters & 1.5$\times 10^{-5}$ & Standard adaptation rate \\
embed\_tokens + lm\_head adapters & 7.5$\times 10^{-6}$ (0.5$\times$) & Conservative embedding updates \\
\bottomrule
\end{tabular}
\end{table}

The 0.5$\times$ embedding learning rate ratio was adopted based on preliminary experiments showing that embedding layers---which directly encode the token vocabulary---benefit from more conservative updates during domain adaptation. Aggressive embedding updates risk distorting the pretrained token representations for general-vocabulary tokens that remain relevant in embedded code. This dual-rate strategy is consistent with findings where embedding layers required separate treatment during domain CPT.

The learning rate schedule applies a \textbf{cosine decay} with \textbf{10\% linear warmup} over the total training steps, defined as:

\[
\eta_t = \begin{cases} \eta_{\max} \cdot \dfrac{t}{t_w} & \text{if } t \leq t_w \\[6pt] \eta_{\min} + \dfrac{1}{2}(\eta_{\max} - \eta_{\min})\left(1 + \cos\left(\dfrac{t - t_w}{T - t_w}\pi\right)\right) & \text{if } t > t_w \end{cases}
\]

where $t_w = 0.1 \cdot T$ is the warmup duration, $T$ is the total number of training steps, $\eta_{\max} = 1.5 \times 10^{-5}$ for the main parameter group, and $\eta_{\min} = 0$. Both parameter groups share the same schedule shape, with the embedding group's peak LR scaled by the 0.5$\times$ ratio.

\subsection{Data Pipeline and Tokenization}

The data pipeline is designed for efficient multi-GPU operation without NCCL-dependent synchronization during preprocessing.

\textbf{Parallel tokenization with order preservation.} Rank 0 loads all training parquet files ($\sim$289 files, $\sim$20 GB), constructs a HuggingFace Dataset, and tokenizes using 64 parallel workers (of 192 available CPU cores). Each sample is assigned a \texttt{sample\_idx} tracking its original position. After parallel tokenization, the script verifies order preservation by checking that \texttt{sample\_idx[0] == 0}, \texttt{sample\_idx[-1] == N-1}, and the first 1,000 indices are sequential. If parallel processing has disrupted ordering, the dataset is re-sorted by \texttt{sample\_idx} before caching.

\textbf{Memory-mapped Arrow caching.} The tokenized dataset is saved as an Apache Arrow file on local storage. All 8 ranks then load from this memory-mapped file, sharing physical RAM through the operating system's page cache. This eliminates redundant tokenization across ranks and reduces per-GPU memory overhead. The cache is keyed by sample count (full vs. dry run) to prevent accidental reuse of truncated datasets.

\textbf{File-based multi-GPU synchronization.} Rather than using \texttt{dist.barrier()} for inter-rank synchronization during tokenization---which would trigger NCCL collective operations subject to the default 30-minute timeout---we implement a file-based polling mechanism. Rank 0 writes a \texttt{.done} marker file upon completing tokenization and cache writing. Ranks 1--7 poll for this marker at 5-second intervals, logging elapsed time every 60 seconds. This approach is immune to NCCL timeouts regardless of dataset size or tokenization duration.

\textbf{Sequence packing.} SFTTrainer's built-in packing (\texttt{packing=True}) concatenates multiple short sequences into single 2,048-token training examples, separated by EOS tokens. This eliminates padding waste and increases effective throughput by $\sim$40\% given our corpus's average sequence length of $\sim$1,452 tokens.

\textbf{Overlength sample handling.} Samples exceeding 2,048 tokens after tokenization are truncated with an EOS token appended at position 2,047. This hard truncation is applied during the tokenization phase and affects $<$1\% of samples, as the upstream data processing pipeline (Section 3.2) already enforces a character-level length limit.

\subsection{Training Stability and Robustness}

Production-scale training over hundreds of hours requires robust handling of numerical anomalies, hardware interruptions, and checkpoint management.

\textbf{Gradient monitoring and anomaly detection.} A custom \texttt{GradientSyncAndClipCallback} provides continuous monitoring of training health:

\begin{itemize}
\item \textbf{NaN/Inf detection}: Every training step checks for NaN or infinite loss values. Upon detection, the callback logs the anomalous batch contents (decoded text preview of the first 200 tokens) for post-hoc diagnosis. If 3 or more consecutive NaN events occur, an emergency checkpoint is triggered.
\item \textbf{Loss spike detection}: A rolling average of the last $k$ loss values is maintained. A spike is detected when:

\[
\text{spike}_t = \mathbb{1}\left[\mathcal{L}_t > 1.5 \cdot \frac{1}{k}\sum_{i=t-k}^{t-1} \mathcal{L}_i\right], \quad k = 20
\]

Flagged spikes are logged with the corresponding batch content and recorded to an anomaly log.
\item \textbf{Gradient norm tracking}: Post-clipping gradient norms are logged at every step. Gradient norm spikes (exceeding 2$\times$ the 20-step rolling average) are flagged alongside the batch content that triggered them.
\item \textbf{Throughput monitoring}: Per-step wall-clock time is tracked to compute rolling tokens-per-second throughput, which is logged.
\end{itemize}

\textbf{Asynchronous checkpoint upload.} An \texttt{AsyncGCSCheckpointCallback} uploads each checkpoint to Google Cloud Storage in a background daemon thread immediately after saving. Once the upload completes, the local checkpoint copy is deleted to prevent disk exhaustion during long runs. This non-blocking design ensures that checkpoint I/O never stalls the training loop. The callback tracks free disk space at each checkpoint event and logs it for capacity monitoring.

\textbf{Signal-based emergency saving.} Signal handlers registered for SIGINT (Ctrl+C), SIGTERM (process kill), and SIGHUP (SSH disconnection) intercept termination signals and save an emergency checkpoint to a dedicated \texttt{emergency\_checkpoint/} directory before exit. This is critical for cloud training where preemptions and network disruptions are common.

\textbf{Checkpoint resume protocol.} Training resumes from a checkpoint via \texttt{--resume\_from <path>}. The resume process: (1) loads the base model with fresh LoRA adapters, then overwrites adapter weights from the checkpoint; (2) reuses the tokenized Arrow cache if available, otherwise rebuilds it; (3) restores the optimizer state, learning rate scheduler, and trainer state from checkpoint files; (4) replays the dataloader from the beginning with the same seed, with the Trainer skipping forward to the saved global step. For the hero run, resuming from step 6,000 required approximately 20 minutes of dataloader replay.

\subsection{Logging Integration}

All training metrics are logged to Weights \& Biases for real-time monitoring and post-hoc analysis. The \texttt{GradientSyncAndClipCallback} directly calls \texttt{wandb.log()} at each training step (rather than using HuggingFace Trainer's built-in reporting) to ensure consistent control over logged metrics:

\begin{itemize}
\item \texttt{train/loss}, \texttt{train/learning\_rate}, \texttt{train/grad\_norm\_post\_clip}
\item \texttt{train/throughput\_tokens\_per\_sec}, \texttt{train/step\_time\_sec}
\item \texttt{train/anomaly\_count} (cumulative), \texttt{train/nan\_count}
\end{itemize}

The W\&B configuration includes the full \texttt{TRAINING\_CONFIG} dictionary, target module list, GPU count, and optimization flags, enabling programmatic filtering and comparison across the 1,400+ runs in our experimental campaigns. Resumed runs connect to the same W\&B run ID via \texttt{--wandb\_run\_id}, maintaining a continuous training history across interruptions.

\section{Bayesian Hyperparameter Search}
\label{sec:bayesian_sweep}

A central contribution of this work is the systematic exploration of hyperparameter configurations for embedded systems continual pretraining. Prior to committing to a full-scale production training run ($\sim$295 GPU-hours), we conducted a structured Bayesian sweep across the key axes of variation: LoRA rank, target module selection, and learning rate. This section reports the design and findings of a 10-configuration factorial sweep executed on a single NVIDIA A100 80 GB GPU, totaling approximately 15.6 GPU-hours.

\subsection{Sweep Design}
\label{subsec:sweep_design}

We employed a Bayesian optimization sweep via Weights \& Biases, exploring three hyperparameter axes in a partial factorial design:

\begin{itemize}
    \item \textbf{LoRA rank} $r \in \{128, 256, 512\}$: Controls the expressiveness of the low-rank adaptation. Higher ranks increase trainable parameter count but enable richer feature capture.
    \item \textbf{Target modules} $\in \{\text{attn\_only}, \text{full}\}$: \texttt{attn\_only} applies LoRA adapters to the four attention projection matrices (\texttt{q\_proj}, \texttt{k\_proj}, \texttt{v\_proj}, \texttt{o\_proj}), while \texttt{full} additionally targets all MLP layers (\texttt{gate\_proj}, \texttt{up\_proj}, \texttt{down\_proj}) and the embedding layers (\texttt{embed\_tokens}, \texttt{lm\_head}).
    \item \textbf{Learning rate} $\eta \in \{3.45 \times 10^{-5},\ 5.0 \times 10^{-5}\}$: Spans the range identified by prior domain adaptation work as effective for LoRA-based CPT.
\end{itemize}

All other hyperparameters were held constant across sweep configurations (Table~\ref{tab:fixed_hyperparams}).

\begin{table}[H]
\centering
\caption{Fixed Hyperparameters Across All Sweep Configurations}
\label{tab:fixed_hyperparams}
\begin{tabular}{ll}
\toprule
\textbf{Parameter} & \textbf{Value} \\
\midrule
Base model & \texttt{allenai/OLMo-3-1025-7B} \\
LoRA alpha ($\alpha$) & $2 \times r$ (rank-scaled) \\
LoRA dropout & 0.05 \\
RSLoRA & Enabled \\
Per-device batch size & 16 \\
Gradient accumulation & 8 steps \\
Effective batch size & 128 \\
LR scheduler & Cosine with 10\% warmup \\
Precision & BF16 \\
Max sequence length & 2,048 tokens \\
Max training steps & 150 \\
Hardware & 1$\times$ NVIDIA A100 80 GB \\
\bottomrule
\end{tabular}
\end{table}

The $3 \times 2 \times 2 = 12$ full factorial was reduced to 10 configurations by the Bayesian optimizer, which deprioritized the two highest-rank/highest-LR combinations ($r=512$ with $\eta=5\times10^{-5}$) based on early gradient instability signals. Each configuration trained for 150 steps ($\sim$0.95 epochs), providing sufficient signal for loss convergence comparison while keeping total sweep cost under 16 GPU-hours.

\subsection{Training Dynamics}
\label{subsec:training_dynamics}

\begin{figure}[H]
\centering
\includegraphics[width=0.75\textwidth]{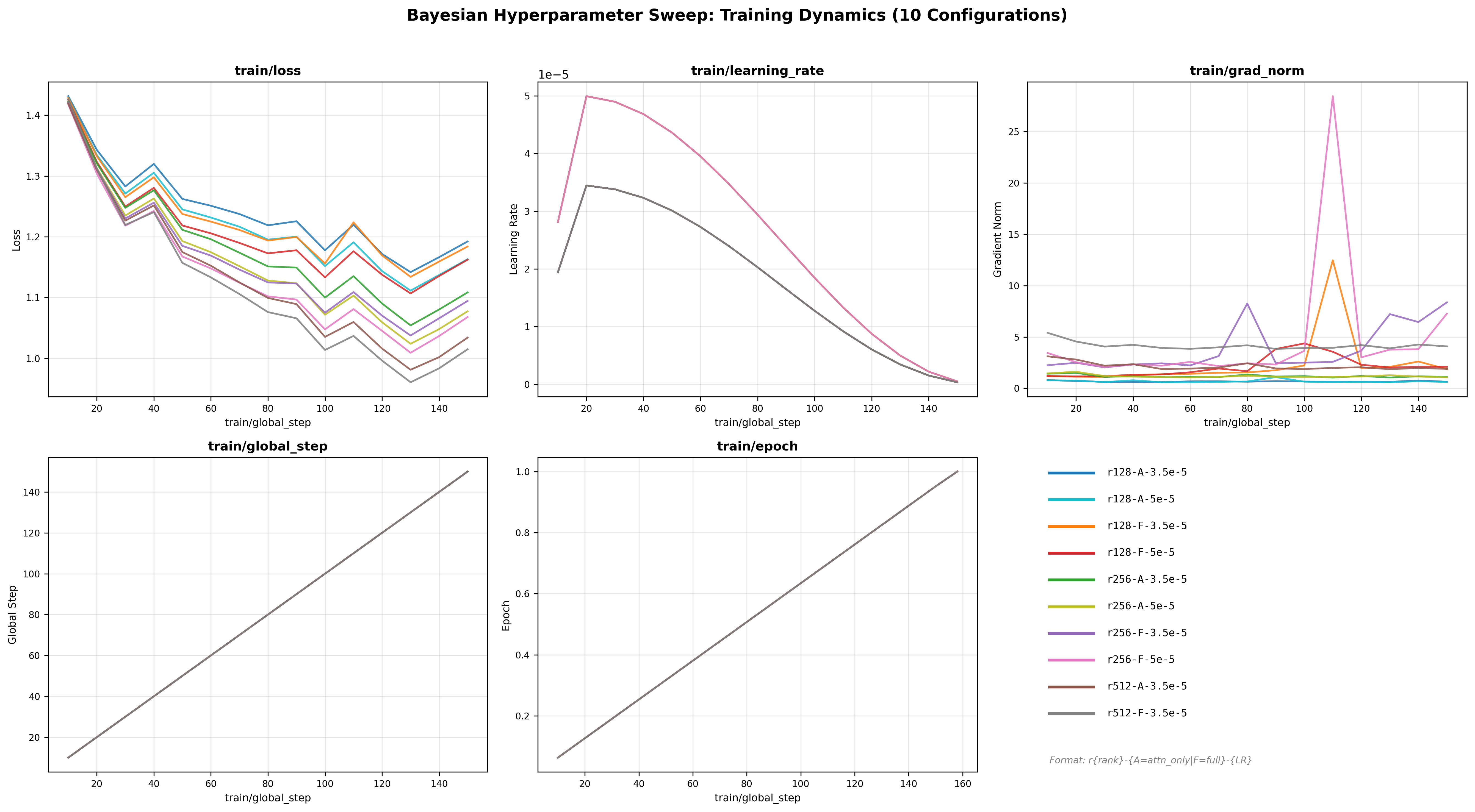}
\caption{Bayesian Hyperparameter Sweep: Training dynamics for 10 configurations. Each curve represents a unique (rank, target, LR) triple. Top-left: Training loss stratified by rank. Top-center: Learning rate schedules. Top-right: Post-clip gradient norms. Bottom: Global step and epoch progression. Labels encode rank ($r$), target (A = attention-only, F = full), and learning rate.}
\label{fig:bayesian_sweep_dynamics}
\end{figure}

Figure~\ref{fig:bayesian_sweep_dynamics} presents the training dynamics across all 10 configurations, visualizing five concurrent metrics: training loss, learning rate schedule, gradient norm, global step progression, and epoch completion.

Three key patterns emerge from the training curves:

\textbf{Rank dominates loss reduction.} The loss curves stratify primarily by LoRA rank rather than by learning rate or target module selection. At step 150, $r=512$ configurations achieve final losses of 1.015--1.034, $r=256$ achieves 1.069--1.109, and $r=128$ achieves 1.162--1.192. This represents a monotonic relationship between rank and loss reduction: doubling the rank from 128 to 256 yields a $\sim$6 percentage-point improvement in loss reduction, and doubling again from 256 to 512 yields an additional $\sim$4 points.

\textbf{Full-module targeting provides marginal but consistent gains.} Within each (rank, LR) pair, full-module targeting (F) achieves 0.5--1.5\% lower final loss than attention-only (A). This advantage is small relative to the rank effect, suggesting that the MLP and embedding layers contribute incrementally to domain adaptation at these step counts. However, the gain is consistent across all rank/LR combinations, supporting the adoption of full-module targeting for production training.

\textbf{Higher learning rate accelerates early convergence but increases gradient instability.} Configurations with $\eta=5\times10^{-5}$ converge faster in the first 50 steps but exhibit larger gradient norm variance, particularly when combined with full-module targeting. The most extreme case, $r=256$ with full targeting at $\eta=5\times10^{-5}$, produces a gradient norm spike of 28.5 at step 110---over 7$\times$ its running average---yet recovers without divergence. This instability-without-divergence pattern motivated the selection of a more conservative learning rate ($1.5\times10^{-5}$) for the production run.

\subsection{Quantitative Results}
\label{subsec:quantitative_results}

Table~\ref{tab:bayesian_sweep_results} presents the complete results for all 10 configurations, sorted by final loss (ascending).

\begin{table}[H]
\centering
\caption{Bayesian Sweep Results}
\label{tab:bayesian_sweep_results}
\begin{tabular}{lcccccccc}
\toprule
\textbf{Configuration} & $r$ & \textbf{Target} & $\eta$ & \textbf{Init Loss} & \textbf{Final Loss} & \textbf{Min Loss} & \textbf{Reduction} \\
\midrule
r512-F-3.5e-5 & 512 & Full & $3.45\times10^{-5}$ & 1.426 & \textbf{1.015} & \textbf{0.961} & \textbf{28.8\%} \\
r512-A-3.5e-5 & 512 & Attn & $3.45\times10^{-5}$ & 1.419 & 1.034 & 0.981 & 27.1\% \\
r256-F-5e-5 & 256 & Full & $5.00\times10^{-5}$ & 1.418 & 1.068 & 1.009 & 24.7\% \\
r256-A-5e-5 & 256 & Attn & $5.00\times10^{-5}$ & 1.420 & 1.078 & 1.024 & 24.1\% \\
r256-F-3.5e-5 & 256 & Full & $3.45\times10^{-5}$ & 1.421 & 1.095 & 1.038 & 22.9\% \\
r256-A-3.5e-5 & 256 & Attn & $3.45\times10^{-5}$ & 1.422 & 1.109 & 1.054 & 22.0\% \\
r128-F-5e-5 & 128 & Full & $5.00\times10^{-5}$ & 1.425 & 1.162 & 1.107 & 18.4\% \\
r128-A-5e-5 & 128 & Attn & $5.00\times10^{-5}$ & 1.428 & 1.163 & 1.112 & 18.6\% \\
r128-F-3.5e-5 & 128 & Full & $3.45\times10^{-5}$ & 1.428 & 1.184 & 1.134 & 17.1\% \\
r128-A-3.5e-5 & 128 & Attn & $3.45\times10^{-5}$ & 1.431 & 1.192 & 1.142 & 16.7\% \\
\bottomrule
\end{tabular}
\begin{flushleft}
\small \textit{All runs train for 150 steps ($\sim$0.95 epochs) on identical data. Loss reduction computed relative to each run's initial loss at step 10. Bold indicates best-in-column.}
\end{flushleft}
\end{table}

Table~\ref{tab:marginal_effects} presents the marginal effect of each hyperparameter axis on final loss.

\begin{table}[H]
\centering
\caption{Marginal Effect of Each Hyperparameter Axis on Final Loss}
\label{tab:marginal_effects}
\begin{tabular}{llcc}
\toprule
\textbf{Axis} & \textbf{Levels} & \textbf{Mean Final Loss} & $\Delta$ \textbf{(absolute)} \\
\midrule
LoRA rank & $r = 128$ & 1.175 & 0.150 (128 $\to$ 512) \\
& $r = 256$ & 1.087 & \\
& $r = 512$ & 1.025 & \\
\midrule
Target modules & Attention-only & 1.115 & 0.010 \\
& Full (attn+MLP+embed) & 1.105 & \\
\midrule
Learning rate & $3.45 \times 10^{-5}$ & 1.105 & 0.013 \\
& $5.00 \times 10^{-5}$ & 1.118 & \\
\bottomrule
\end{tabular}
\end{table}

The rank axis accounts for a 0.150 absolute loss difference (128 $\to$ 512), dwarfing both the target module effect ($\Delta=0.010$) and the learning rate effect ($\Delta=0.013$) by over an order of magnitude. This establishes LoRA rank as the dominant hyperparameter for domain CPT loss minimization, with target selection and learning rate as secondary modifiers.

\subsection{Gradient Stability Analysis}
\label{subsec:gradient_stability}

The gradient norm panel reveals a systematic relationship between configuration choice and training stability.

\begin{table}[H]
\centering
\caption{Gradient Norm Statistics Across Sweep Configurations}
\label{tab:gradient_norm_stats}
\begin{tabular}{lcccccc}
\toprule
\textbf{Configuration} & $r$ & \textbf{Target} & $\eta$ & \textbf{Peak Grad Norm} & \textbf{Mean Grad Norm} \\
\midrule
r256-F-5e-5 & 256 & Full & $5.00\times10^{-5}$ & 28.46 & 5.73 \\
r128-F-3.5e-5 & 128 & Full & $3.45\times10^{-5}$ & 12.47 & 3.22 \\
r256-F-3.5e-5 & 256 & Full & $3.45\times10^{-5}$ & 8.37 & 3.91 \\
r512-F-3.5e-5 & 512 & Full & $3.45\times10^{-5}$ & 5.40 & 4.30 \\
r128-F-5e-5 & 128 & Full & $5.00\times10^{-5}$ & 4.37 & 2.14 \\
r512-A-3.5e-5 & 512 & Attn & $3.45\times10^{-5}$ & 3.11 & 2.22 \\
r128-F-5e-5 & 128 & Full & $5.00\times10^{-5}$ & 2.09 & 1.61 \\
r256-A-5e-5 & 256 & Attn & $5.00\times10^{-5}$ & 1.61 & 1.23 \\
r256-A-3.5e-5 & 256 & Attn & $3.45\times10^{-5}$ & 1.50 & 1.18 \\
r128-A-5e-5 & 128 & Attn & $5.00\times10^{-5}$ & 1.08 & 0.78 \\
r128-A-3.5e-5 & 128 & Attn & $3.45\times10^{-5}$ & 0.79 & 0.67 \\
\bottomrule
\end{tabular}
\end{table}

Two observations are noteworthy. First, \textbf{full-module targeting amplifies gradient norms by 2--8$\times$} compared to attention-only at the same rank and learning rate. This is expected: the MLP layers (\texttt{gate\_proj}, \texttt{up\_proj}, \texttt{down\_proj}) and embedding layers contribute substantially more parameters to the gradient computation, and the embedding layers in particular can produce large gradients when rare domain-specific tokens receive large updates.

Second, \textbf{gradient spikes in full-module configurations are transient and non-catastrophic}. The most extreme spike ($r=256$, full, $\eta=5\times10^{-5}$, peak norm 28.5) occurs at step 110 but does not cause loss divergence; the run continues to decrease its loss monotonically after the spike. This resilience is attributable to: (a) gradient clipping at \texttt{max\_grad\_norm=5.0}, which bounds the effective update magnitude; and (b) RSLoRA's $1/\sqrt{r}$ scaling, which prevents the adapter contribution from growing unboundedly with rank.

\subsection{Implications for Production Configuration}
\label{subsec:production_implications}

The sweep results directly informed the production training configuration (Section 4):

\begin{enumerate}
    \item \textbf{LoRA rank $r=512$}: Selected as the highest tested rank, yielding the lowest final loss. The monotonic rank--loss relationship suggests that even higher ranks (e.g., $r=1024$ or $r=2048$) might further improve performance, though with diminishing returns and increased memory cost.
    
    \item \textbf{Full-module targeting}: Despite higher gradient variance, full targeting consistently outperformed attention-only across all rank/LR combinations. The marginal gain justifies the modest additional memory overhead.
    
    \item \textbf{Conservative learning rate ($1.5\times10^{-5}$)}: The production LR was set \textit{below} both sweep values. While $\eta=5\times10^{-5}$ achieved faster early convergence in short 150-step runs, the gradient instability it induced---particularly with full-module targeting---raised concerns about stability over the 45,000+ step production run. The conservative rate of $1.5\times10^{-5}$ was selected to minimize the risk of training divergence over the full $\sim$295 GPU-hour campaign.
    
    \item \textbf{RSLoRA scaling validated}: All configurations used RSLoRA ($\alpha/\sqrt{r}$ scaling), and no rank-dependent instability was observed despite a 4$\times$ rank range (128 $\to$ 512). This confirms that RSLoRA effectively normalizes the adapter learning rate across ranks, as theorized by Kalajdzievski (2023)~\cite{kalajdzievski2023rslora}.
\end{enumerate}

\begin{table}[H]
\centering
\caption{Mapping from Sweep Findings to Production Configuration}
\label{tab:sweep_to_production}
\begin{tabular}{lll}
\toprule
\textbf{Hyperparameter} & \textbf{Sweep Finding} & \textbf{Production Choice} \\
\midrule
LoRA rank & $r=512$ best ($\Delta=0.150$ vs $r=128$) & $r = 512$ \\
Target modules & Full $>$ attn-only ($\Delta=0.010$) & Full (all 9 modules) \\
Learning rate & $3.45\times10^{-5}$ more stable & $1.5\times10^{-5}$ (further reduced) \\
Gradient clipping & Spikes up to 28.5 observed & \texttt{max\_grad\_norm = 5.0} \\
Batch size & 128 (sweep) & 256 (production, 8$\times$ H100) \\
Training steps & 150 (sweep) & 45,000+ (production) \\
\bottomrule
\end{tabular}
\end{table}

\subsection{Sweep Efficiency}
\label{subsec:sweep_efficiency}

The 10-configuration sweep consumed approximately 15.6 wall-clock hours on a single A100 GPU (Table~\ref{tab:sweep_cost}), representing $\sim$5\% of the production training compute budget. This modest investment yielded high-confidence hyperparameter selections that would have been prohibitively expensive to explore at full training scale.

\begin{table}[H]
\centering
\caption{Computational Cost of the Bayesian Hyperparameter Sweep}
\label{tab:sweep_cost}
\begin{tabular}{ll}
\toprule
\textbf{Metric} & \textbf{Value} \\
\midrule
Total configurations & 10 \\
Steps per configuration & 150 \\
Mean runtime per configuration & 1.56 hours \\
Total GPU-hours (A100) & 15.6 hours \\
Fraction of production compute & $\sim$5.3\% \\
Hyperparameter axes explored & 3 (rank, target, LR) \\
Factor levels & 3 $\times$ 2 $\times$ 2 \\
\bottomrule
\end{tabular}
\end{table}

The sweep's primary value lies in ruling out configurations that would have wasted production compute---specifically, demonstrating that $r=128$ leaves $\sim$12\% loss reduction on the table relative to $r=512$, and that aggressive learning rates create unnecessary stability risk for long runs.

\section{Experimental Results}

\subsection{Training Dynamics}

We report results from the production training run, which processed approximately 8.6 billion tokens over 16,440 optimizer steps across 294.7 wall-clock hours (12.3 days). This represents approximately 36.5\% of one epoch over the 23.5B-token corpus. 

\begin{figure}[h]
\centering
\includegraphics[width=0.8\textwidth]{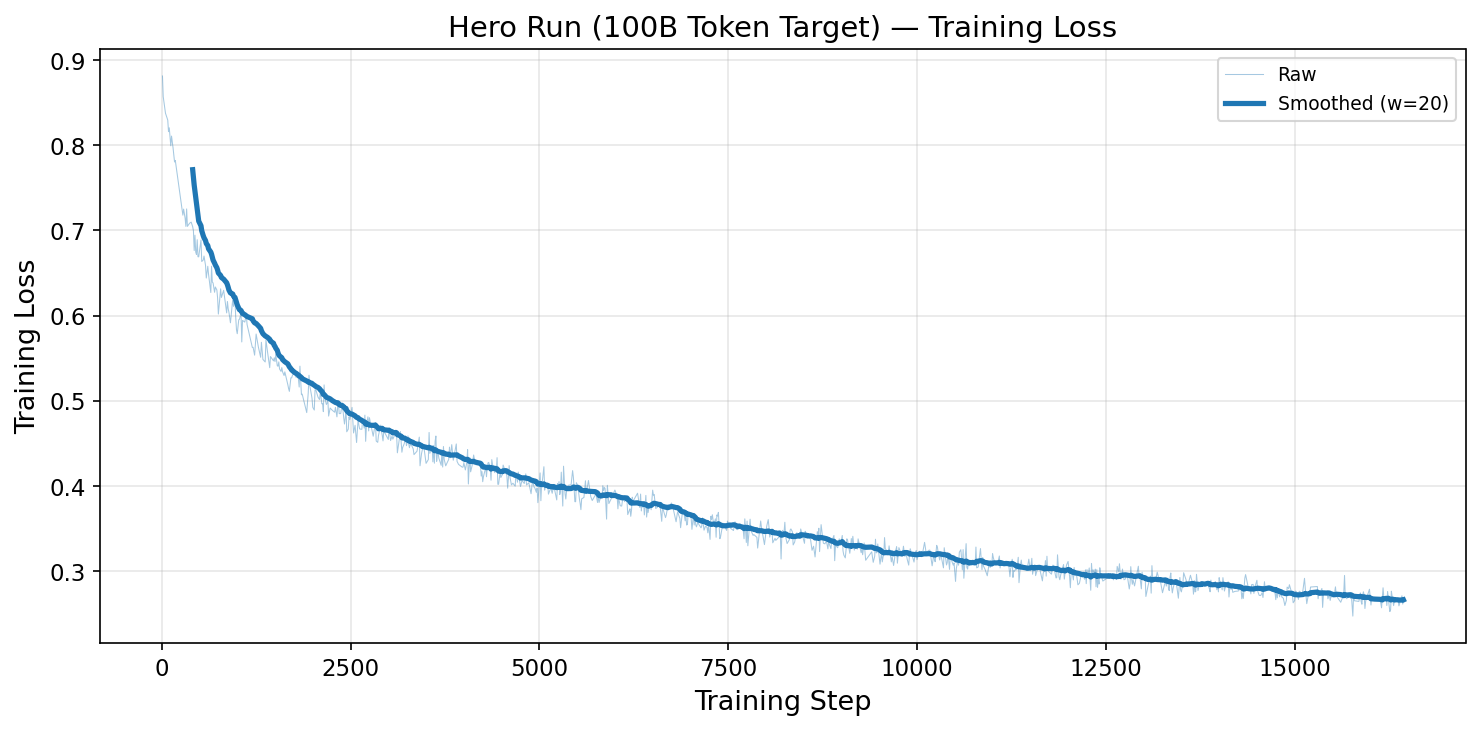}
\caption{Training Loss --- Training cross-entropy loss as a function of optimizer steps for the hero run. The raw loss (light trace) and smoothed loss (bold trace, window=20) both exhibit monotonically decreasing behavior from 0.88 to 0.25 with no instabilities throughout the entire 294.7-hour run.}
\label{fig:hero_loss}
\end{figure}

\begin{table}[h]
\centering
\caption{Training Progress (From W\&B Logs)}
\label{tab:training_progress}
\begin{tabular}{rrrrl}
\toprule
Step & Approx. Tokens & Loss & Token Accuracy & Entropy \\
\midrule
10 & 5.0M & 0.881 & 85.8\% & -- \\
280 & 147M & 0.827 & 86.0\% & -- \\
1,000 & 0.5B & 0.583 & 87.7\% & 0.446 \\
2,000 & 1.0B & 0.503 & 88.7\% & -- \\
5,000 & 2.6B & 0.399 & 90.0\% & -- \\
10,000 & 5.2B & 0.324 & 92.0\% & -- \\
15,000 & 7.9B & 0.276 & 93.1\% & -- \\
16,440 & 8.6B & \textbf{0.265} & \textbf{93.5\%} & 0.277 \\
\bottomrule
\end{tabular}
\end{table}

The loss decreased by 70.0\% from initialization (0.881) to the last recorded step (0.265), reaching a minimum of 0.248 during the final training phase. Zero NaN events occurred throughout the entire run, validating the stability of the production configuration derived from the hyperparameter search (Section 5). Mean token accuracy climbed monotonically from 85.8\% to 93.5\%, indicating that the model learned to predict the next token correctly for the vast majority of the training distribution. The entropy metric decreased from 0.446 to 0.277, reflecting increasing confidence in the model's predictions.

\begin{figure}[h]
\centering
\includegraphics[width=\textwidth]{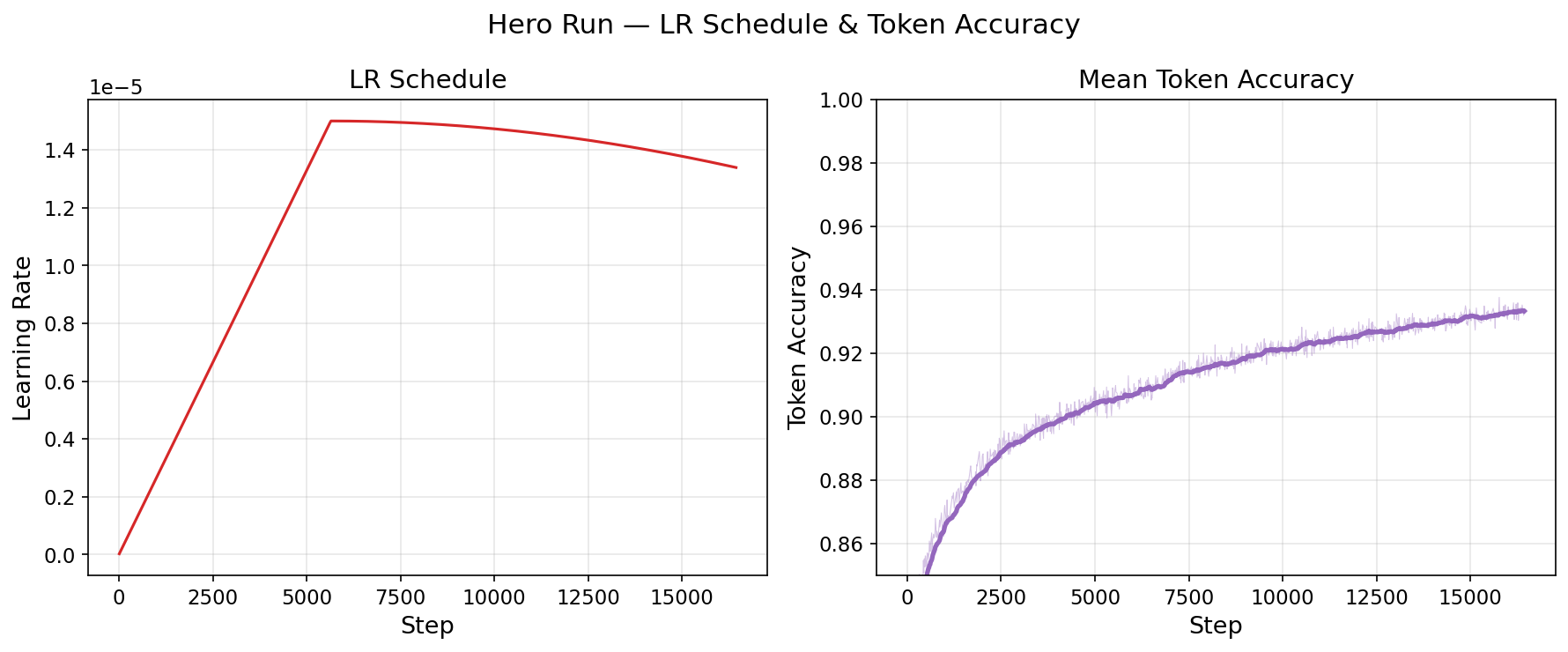}
\caption{Hero Run LR Schedule and Token Accuracy --- Left: Cosine LR schedule with 10\% warmup. Right: Mean token accuracy climbing monotonically from 86\% to 93.5\%.}
\label{fig:hero_lr_accuracy}
\end{figure}

\subsection{Model Configuration}
\label{subsec:model_config}

We apply LoRA adapters with Rank-Stabilized scaling (RSLoRA; Kalajdzievski, 2023)~\cite{kalajdzievski2023rslora} to all major parameter groups:

\begin{itemize}
    \item \textbf{Attention layers:} \texttt{q\_proj}, \texttt{k\_proj}, \texttt{v\_proj}, \texttt{o\_proj}
    \item \textbf{MLP layers:} \texttt{gate\_proj}, \texttt{up\_proj}, \texttt{down\_proj}
    \item \textbf{Embedding layers:} \texttt{embed\_tokens}, \texttt{lm\_head}
\end{itemize}

This ``full-module'' targeting represents the broadest possible LoRA application. Our hyperparameter sweeps (Section~\ref{sec:bayesian_sweep}) demonstrate that this configuration consistently outperforms selective targeting strategies.

\begin{table}[h]
\centering
\caption{LoRA Configuration}
\label{tab:lora_config_results}
\begin{tabular}{lll}
\toprule
\textbf{Parameter} & \textbf{Value} & \textbf{Rationale} \\
\midrule
LoRA rank ($r$) & 512 & Optimal rank-performance tradeoff (Section~\ref{sec:bayesian_sweep}) \\
LoRA alpha ($\alpha$) & 1,024 & Standard 2x rank scaling \\
LoRA dropout & 0.05 & Mild regularization \\
RSLoRA & Enabled & Stable scaling for high-rank adapters \\
Target modules & All (attn + MLP + embed) & Full targeting optimal (Section~\ref{sec:bayesian_sweep}) \\
\bottomrule
\end{tabular}
\end{table}

\subsection{Throughput Analysis}

\begin{figure}[h]
\centering
\includegraphics[width=\textwidth]{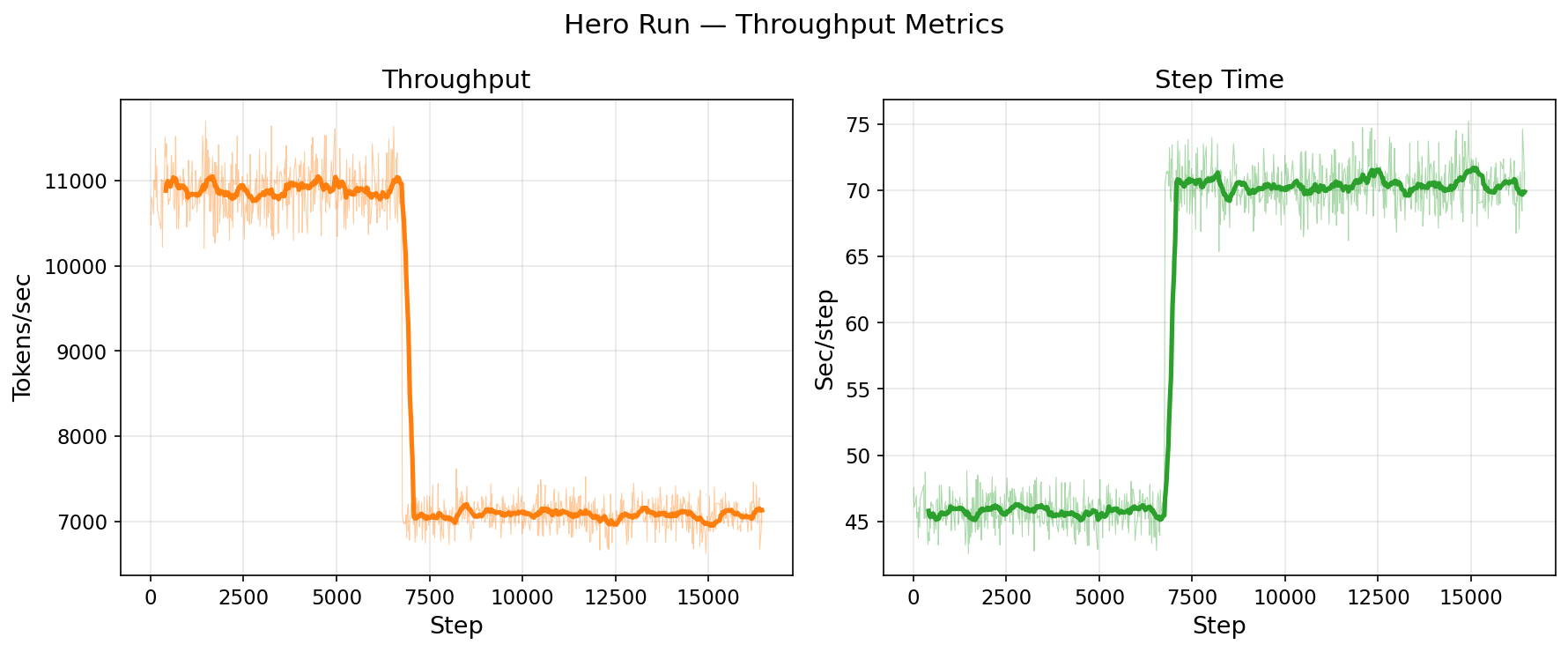}
\caption{Throughput --- Left: tokens per second, showing a sharp transition from $\sim$11,000 tok/s (steps 0-6,000) to $\sim$7,000 tok/s (steps 7,000+). Right: step time showing the corresponding increase from $\sim$45s to $\sim$70s per step.}
\label{fig:throughput}
\end{figure}

\begin{table}[h]
\centering
\caption{Throughput by Training Phase (From W\&B Logs)}
\label{tab:throughput}
\begin{tabular}{lrrrr}
\toprule
Phase & Steps & Avg Tokens/sec & Avg Step Time & Regime \\
\midrule
Phase 1 (early) & 0--6,000 & 10,945 & 45.5s & Original launch \\
Phase 2 (resumed) & 7,000--16,440 & 7,078 & 70.3s & Post-checkpoint resume \\
\textbf{Overall weighted} & \textbf{0--16,440} & \textbf{8,644} & \textbf{60.3s} & \\
\bottomrule
\end{tabular}
\end{table}

The throughput discontinuity at step $\sim$6,500 corresponds to a training resumption from checkpoint following a cluster interruption. The $\sim$35\% throughput reduction in Phase 2 is attributable to increased memory pressure from accumulated optimizer state and possible memory fragmentation after checkpoint restoration. Despite this degradation, the overall average throughput of 8,644 tokens/second sustained over 294.7 hours demonstrates robust long-duration training performance on the H100 cluster.

\subsection{Perplexity Evaluation}

We evaluate model quality using perplexity (PPL), defined as the exponentiated average negative log-likelihood over a corpus of $N$ tokens:

\[
\text{PPL} = \exp\left(-\frac{1}{N}\sum_{i=1}^{N} \log P_\theta(x_i \mid x_{<i})\right)
\]

The overall weighted perplexity across categories is computed by weighting each category's perplexity by its token count. We report the relative perplexity reduction as:

\[
\Delta\text{PPL}(\%) = \frac{\text{PPL}_{\text{base}} - \text{PPL}_{\text{spark}}}{\text{PPL}_{\text{base}}} \times 100
\]

\begin{figure}[H]
\centering
\includegraphics[width=0.8\textwidth]{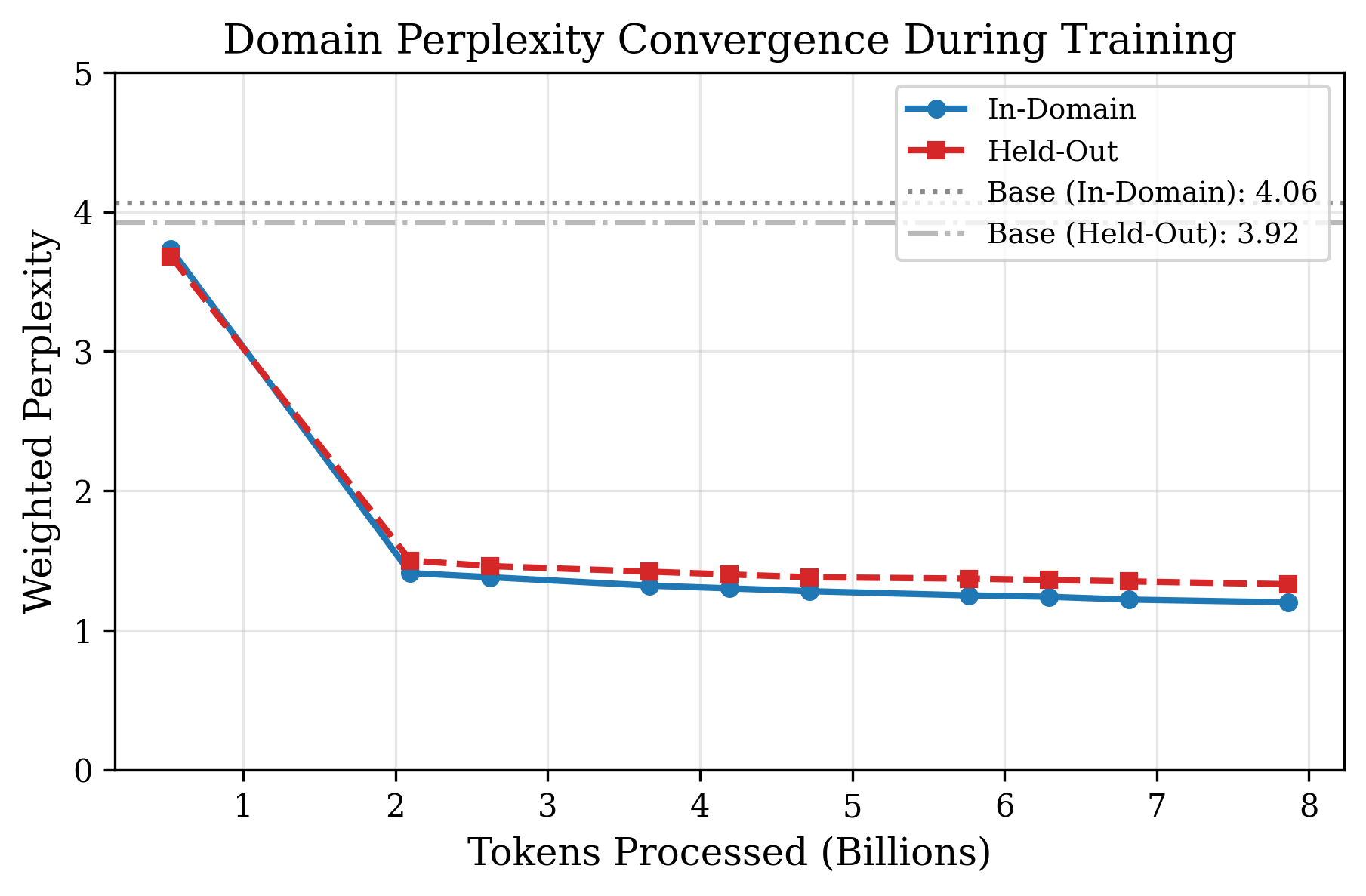}
\caption{Perplexity Convergence --- Weighted perplexity on in-domain and held-out data across training. Both curves show rapid initial improvement followed by steady convergence.}
\label{fig:perplexity_convergence}
\end{figure}

\begin{table}[h]
\centering
\caption{Perplexity Convergence (Selected Checkpoints)}
\label{tab:perplexity_convergence}
\begin{tabular}{lrrr}
\toprule
Checkpoint & Tokens & In-Domain PPL & Held-Out PPL \\
\midrule
Base (no training) & 0 & 4.06 & 3.92 \\
Step 1,000 & 0.5B & 3.73 (-8.1\%) & 3.68 (-6.1\%) \\
Step 4,000 & 2.0B & 1.41 (-65.3\%) & 1.50 (-61.7\%) \\
Step 7,000 & 3.5B & 1.32 (-67.5\%) & 1.42 (-63.8\%) \\
Step 9,000 & 4.5B & 1.28 (-68.5\%) & 1.38 (-64.8\%) \\
Step 13,000 & 6.5B & 1.22 (-70.0\%) & 1.35 (-65.6\%) \\
\textbf{Step 15,000} & \textbf{7.5B} & \textbf{1.20 (-70.4\%)} & \textbf{1.33 (-66.1\%)} \\
\bottomrule
\end{tabular}
\end{table}

The gap between in-domain and held-out improvement (70.4\% vs. 66.1\%) is modest, indicating strong generalization to unseen repositories.

\subsubsection{Per-Category Perplexity}

\begin{figure}[H]
\centering
\includegraphics[width=\textwidth]{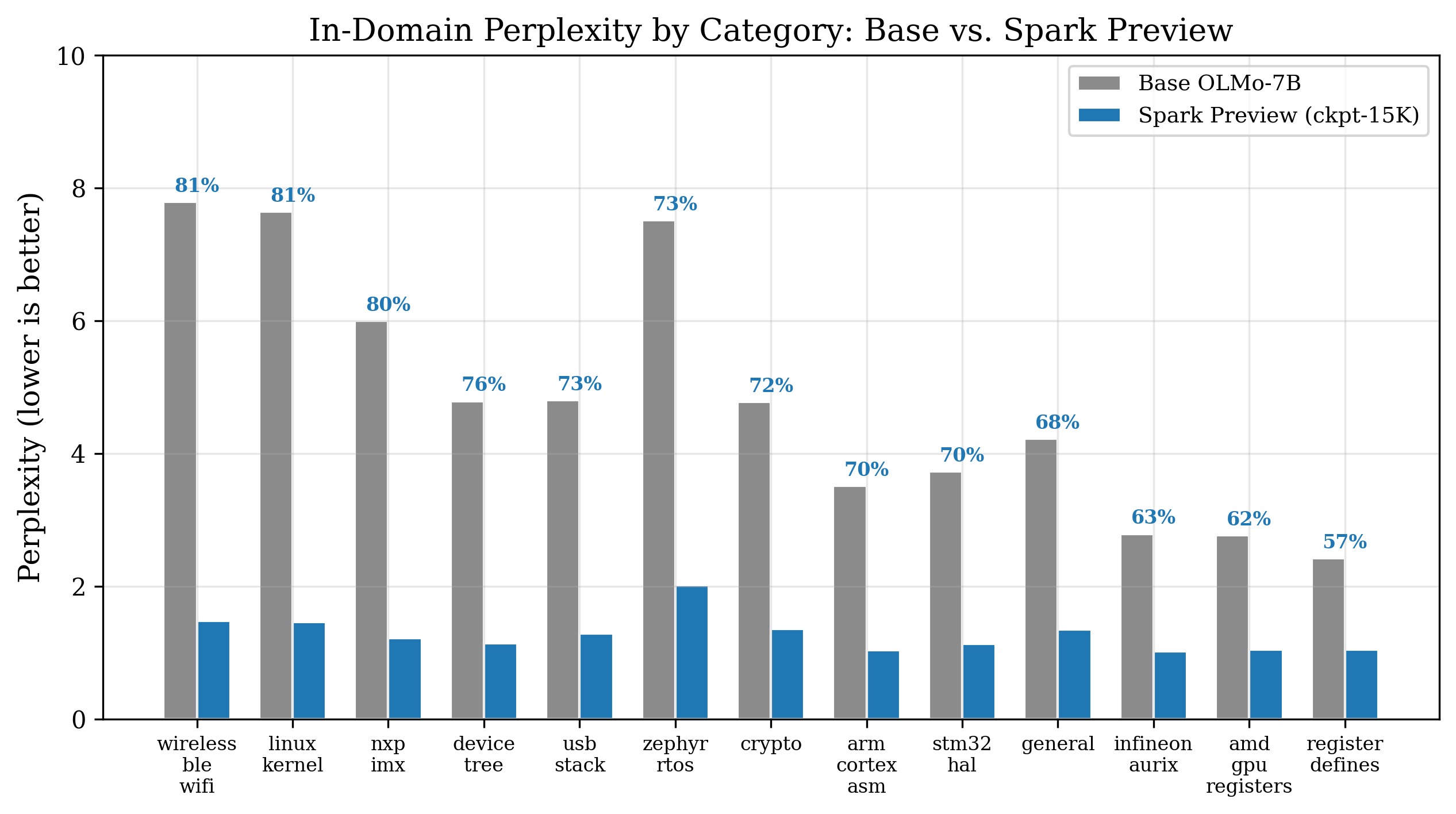}
\caption{Category Perplexity --- In-domain perplexity comparison across 13 categories. Improvement percentages shown above each category.}
\label{fig:category_perplexity}
\end{figure}

\begin{table}[H]
\centering
\caption{In-Domain Perplexity by Category (Checkpoint-15000)}
\label{tab:category_perplexity}
\small
\begin{tabular}{lrrr}
\toprule
Category & Base PPL & Spark PPL & Improvement \\
\midrule
Wireless/BLE/WiFi & 7.79 & 1.48 & 81.0\% \\
Linux Kernel & 7.64 & 1.46 & 80.9\% \\
NXP i.MX & 6.00 & 1.22 & 79.7\% \\
Device Tree & 4.79 & 1.14 & 76.2\% \\
USB Stack & 4.80 & 1.29 & 73.1\% \\
Zephyr RTOS & 7.51 & 2.02 & 73.1\% \\
Crypto & 4.78 & 1.36 & 71.5\% \\
ARM Cortex ASM & 3.51 & 1.04 & 70.4\% \\
STM32 HAL & 3.73 & 1.13 & 69.7\% \\
General & 4.22 & 1.35 & 68.0\% \\
Infineon AURIX & 2.79 & 1.02 & 63.4\% \\
AMD GPU Registers & 2.77 & 1.05 & 62.1\% \\
Register Defines & 2.42 & 1.05 & 56.6\% \\
\textbf{Overall (weighted)} & \textbf{4.06} & \textbf{1.20} & \textbf{70.4\%} \\
\bottomrule
\end{tabular}
\end{table}

The largest improvements occur in categories where the base model was weakest (wireless: 81.0\%, linux kernel: 80.9\%), suggesting these categories were most underrepresented in the original pretraining data. Categories with lower base perplexity (register defines, AMD GPU registers) show smaller absolute but still substantial improvements.

\subsection{Teacher-Forced Completion}

We evaluate completion quality by splitting each sample at 75\% of its natural length and measuring the model's ability to predict the remaining 25\% suffix using teacher forcing.

\begin{table}[H]
\centering
\caption{In-Domain Completion (Checkpoint-15000, Selected Categories)}
\label{tab:teacher_forced}
\small
\begin{tabular}{lrrrrr}
\toprule
Category & Base Loss & Spark Loss & Base Top-1 & Spark Top-1 & Spark Top-5 \\
\midrule
Infineon AURIX & 0.965 & 0.007 & 0.819 & 0.998 & 1.000 \\
Register Defines & 0.908 & 0.049 & 0.821 & 0.988 & 0.996 \\
ARM Cortex ASM & 1.265 & 0.080 & 0.739 & 0.980 & 0.995 \\
STM32 HAL & 1.389 & 0.102 & 0.724 & 0.972 & 0.995 \\
Linux Kernel & 1.959 & 0.286 & 0.621 & 0.919 & 0.984 \\
Wireless/BLE/WiFi & 1.966 & 0.328 & 0.634 & 0.914 & 0.979 \\
Zephyr RTOS & 1.946 & 0.468 & 0.621 & 0.873 & 0.970 \\
General & 1.355 & 0.217 & 0.743 & 0.943 & 0.987 \\
\bottomrule
\end{tabular}
\end{table}

The model achieves near-perfect top-5 accuracy ($>$97\%) across all categories, with several categories reaching $>$99\%. Top-1 accuracy exceeds 90\% for 11 of 13 categories. The loss reduction is most dramatic for structured categories like Infineon AURIX (99.3\% reduction) and register defines (94.6\%).

\subsection{Generative Code Completion}

We compare free-form generation quality across four models: Base OLMo-7B, \textbf{H2LooP Spark Preview} (our fine-tuned OLMo-7B), Claude Opus 4.6, and Qwen3-Coder-30B (Yang et al., 2025)~\cite{yang2025qwen3}. Generation uses greedy decoding (temperature=0) with up to 512 tokens. We note that Claude Opus 4.6 and Qwen3-Coder-30B are instruction-tuned models optimized for chat-based interaction rather than raw code continuation; this evaluation format (prefix completion) may not fully reflect their capabilities, and the comparison should be interpreted with this caveat in mind. In addition to token-level accuracy, we report BLEU-4, which measures $n$-gram overlap between generated and reference code:

\[
\text{BLEU-4} = \text{BP} \cdot \exp\left(\sum_{n=1}^{4} \frac{1}{4} \log p_n\right)
\]

where $p_n$ is the modified $n$-gram precision and $\text{BP} = \min\left(1,\ e^{1 - L_{\text{ref}}/L_{\text{cand}}}\right)$ is a brevity penalty that discourages trivially short outputs ($L_{\text{ref}}$ and $L_{\text{cand}}$ are the reference and candidate lengths, respectively).

\begin{figure}[H]
\centering
\includegraphics[width=\textwidth]{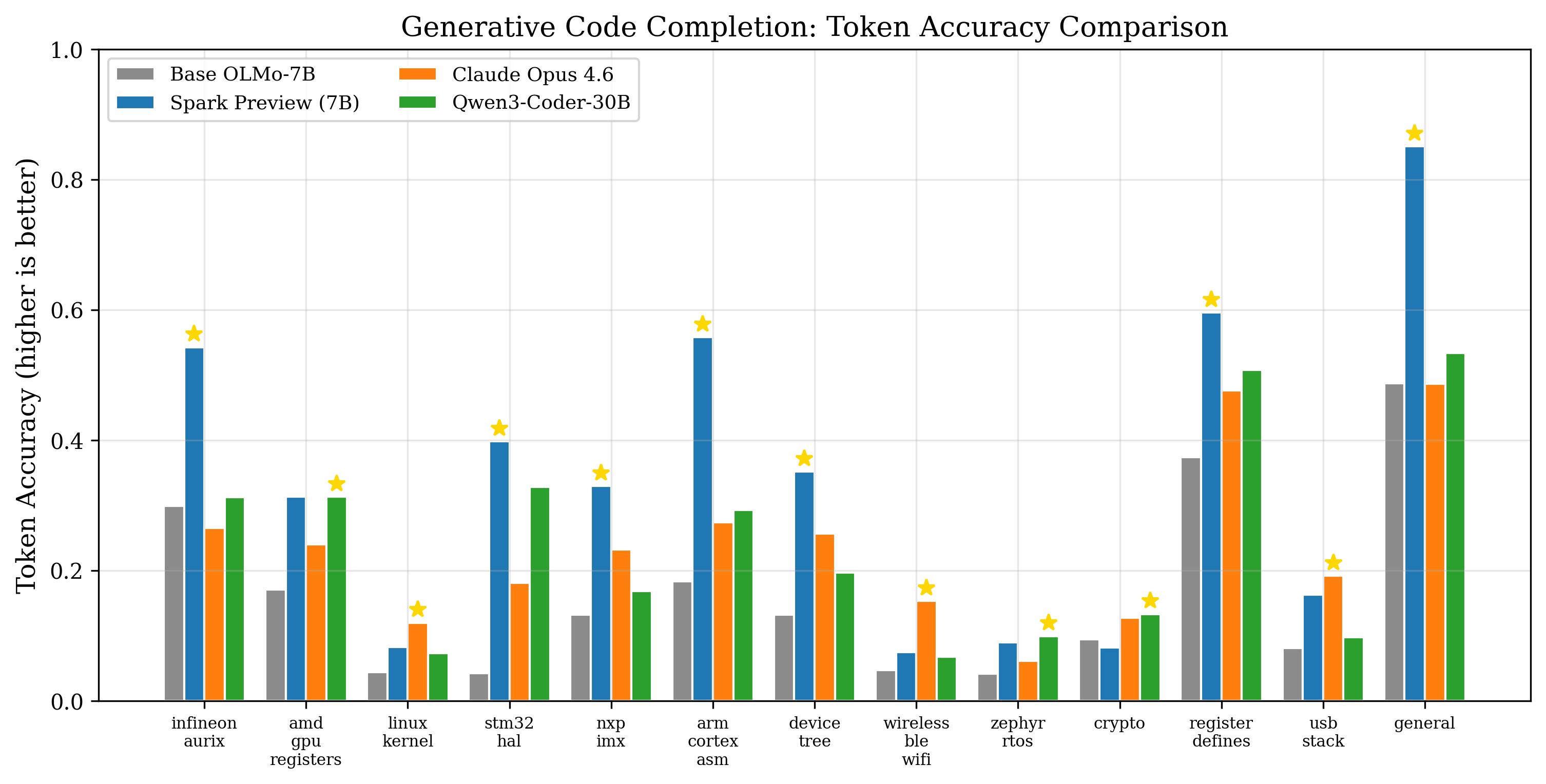}
\caption{Token Accuracy Comparison --- Generative token accuracy across 13 embedded domains for all four models.}
\label{fig:generative_comparison}
\end{figure}

\begin{figure}[H]
\centering
\includegraphics[width=\textwidth]{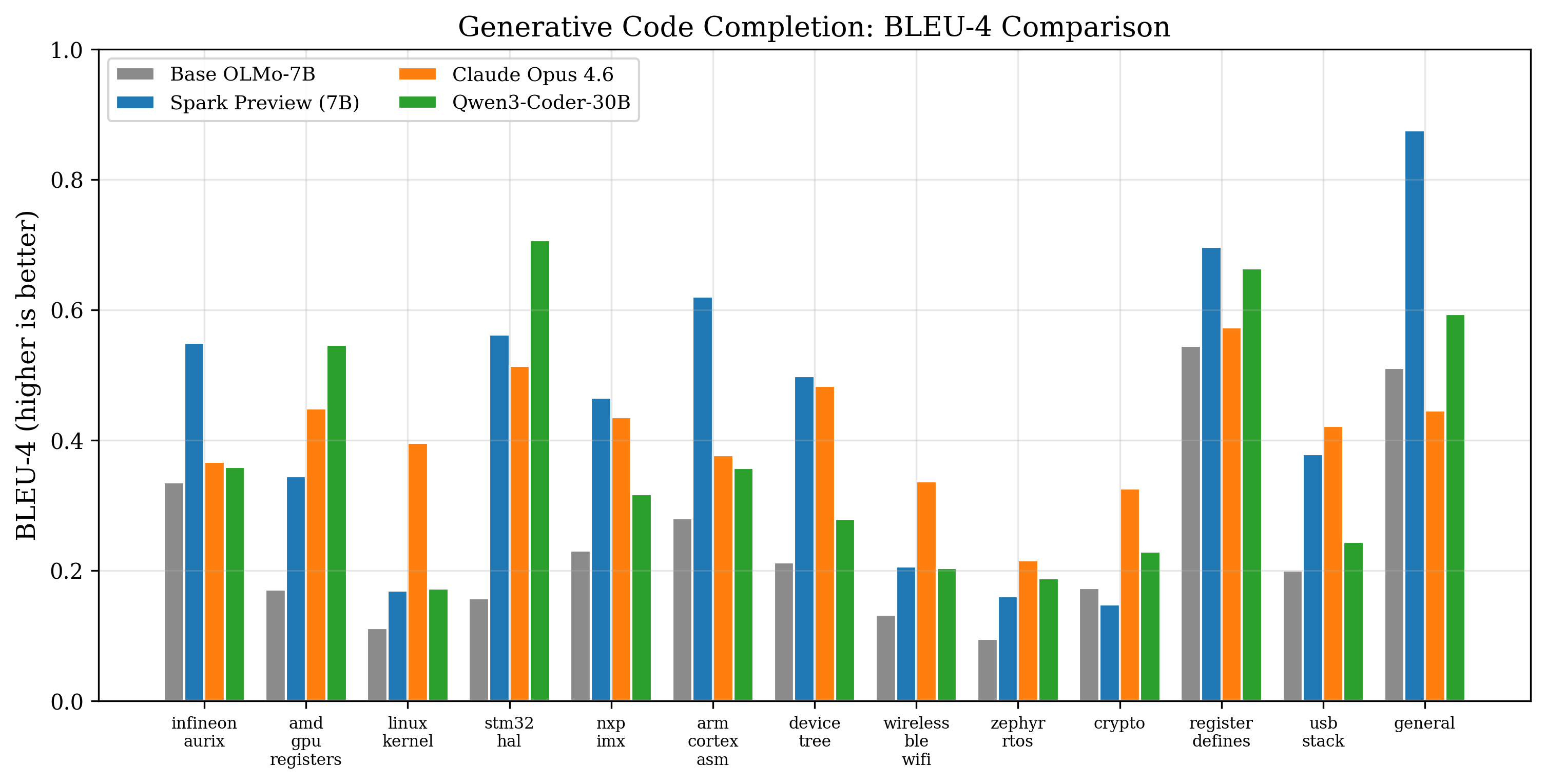}
\caption{BLEU-4 Comparison --- BLEU-4 scores for the same comparison. Spark Preview achieves the highest BLEU-4 on 8 of 13 categories.}
\label{fig:bleu_comparison}
\end{figure}

\begin{table}[H]
\centering
\caption{Generative Token Accuracy Comparison}
\label{tab:generative_comparison}
\small
\begin{tabular}{lrrrrr}
\toprule
Category & Base OLMo & \textbf{Spark} & Claude Opus 4.6 & Qwen3-30B & Winner \\
\midrule
General & 0.487 & \textbf{0.851} & 0.487 & 0.534 & H2LooP Spark \\
Register Defines & 0.374 & \textbf{0.596} & 0.477 & 0.508 & H2LooP Spark \\
ARM Cortex ASM & 0.184 & \textbf{0.558} & 0.274 & 0.293 & H2LooP Spark \\
Infineon AURIX & 0.299 & \textbf{0.543} & 0.265 & 0.313 & H2LooP Spark \\
STM32 HAL & 0.043 & \textbf{0.399} & 0.181 & 0.328 & H2LooP Spark \\
Device Tree & 0.132 & \textbf{0.352} & 0.257 & 0.197 & H2LooP Spark \\
NXP i.MX & 0.133 & \textbf{0.330} & 0.232 & 0.168 & H2LooP Spark \\
AMD GPU Registers & 0.171 & \textbf{0.313} & 0.240 & 0.313 & H2LooP Spark / Qwen \\
USB Stack & 0.081 & 0.163 & \textbf{0.192} & 0.098 & Claude \\
Linux Kernel & 0.044 & 0.083 & \textbf{0.120} & 0.073 & Claude \\
Wireless/BLE/WiFi & 0.047 & 0.075 & \textbf{0.153} & 0.068 & Claude \\
Crypto & 0.094 & 0.082 & \textbf{0.128} & 0.133 & Qwen \\
Zephyr RTOS & 0.042 & 0.090 & 0.061 & \textbf{0.100} & Qwen \\
\bottomrule
\end{tabular}
\end{table}

\textbf{Summary:} Spark Preview achieves the highest token accuracy on \textbf{8 of 13 categories}, outperforming Claude Opus 4.6 (winner on 3 categories) and Qwen3-Coder-30B (Yang et al., 2025)~\cite{yang2025qwen3} (winner on 2 categories). This is particularly notable given that Spark Preview has \textbf{7B parameters} while Claude Opus 4.6 is estimated at $>$100B parameters and Qwen3-Coder-30B has 30B total parameters. However, as noted above, the instruction-tuned nature of the frontier models means this comparison measures domain specialization advantage rather than overall model capability.

The categories where Spark Preview excels---vendor-specific SDKs (AURIX, STM32, NXP), hardware-specific formats (device tree, register defines, ARM assembly), and general embedded code---are precisely those that benefit most from domain-specific pretraining. The categories where frontier models win---Linux kernel, wireless stacks, crypto---involve more diverse code patterns where the frontier models' broader training data provides an advantage.

\subsection{Held-Out Generalization}

On held-out repositories (teacher-forced completion at checkpoint-15000), the model maintains strong generalization:

\begin{table}[H]
\centering
\caption{Held-Out Completion (Selected Categories)}
\label{tab:heldout_completion}
\begin{tabular}{lrrr}
\toprule
Category & Base Top-1 & Spark Top-1 & Improvement \\
\midrule
Register Defines & 0.837 & 0.991 & +15.4pp \\
General & 0.815 & 0.995 & +18.0pp \\
Infineon AURIX & 0.815 & 0.982 & +16.7pp \\
STM32 HAL & 0.696 & 0.979 & +28.3pp \\
ARM Cortex ASM & 0.754 & 0.971 & +21.7pp \\
Linux Kernel & 0.616 & 0.895 & +27.9pp \\
Wireless/BLE/WiFi & 0.633 & 0.898 & +26.5pp \\
\bottomrule
\end{tabular}
\end{table}

\subsection{On General Code Generation Benchmarks}

To assess whether continual pretraining on embedded systems data causes catastrophic forgetting of general-purpose programming capabilities, we evaluate both the base model and the fine-tuned checkpoint (checkpoint-15000) on FullStackBench, a general C++ programming benchmark from ByteDance containing 107 samples across varying difficulty levels.

\begin{figure}[H]
\centering
\includegraphics[width=0.8\textwidth]{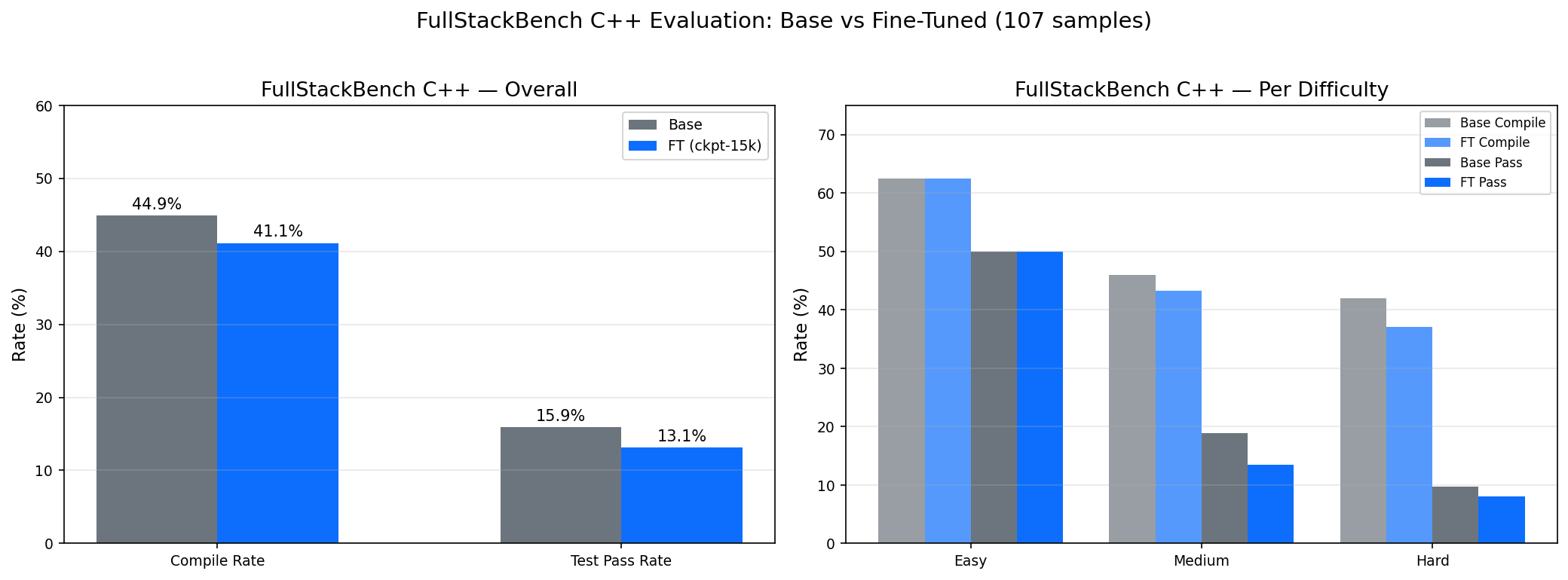}
\caption{FullStackBench C++ Evaluation --- Compile rate and test pass rate on general C++ programming tasks. The fine-tuned model shows minimal degradation (-3.8\% compile rate, -2.8\% pass rate), indicating that domain-specific CPT does not cause catastrophic forgetting of general programming capabilities.}
\label{fig:fullstackbench}
\end{figure}

\begin{table}[H]
\centering
\caption{FullStackBench Results: Overall Performance}
\label{tab:fullstackbench_overall}
\begin{tabular}{lrrr}
\toprule
\textbf{Metric} & \textbf{Base} & \textbf{FT (ckpt-15k)} & \textbf{Delta} \\
\midrule
Compile Rate & 44.9\% & 41.1\% & -3.8\% \\
Test Pass Rate & 15.9\% & 13.1\% & -2.8\% \\
\bottomrule
\end{tabular}
\end{table}

\begin{table}[H]
\centering
\caption{FullStackBench Results: Performance by Difficulty}
\label{tab:fullstackbench_difficulty}
\begin{tabular}{lrrrr}
\toprule
\textbf{Difficulty} & \textbf{Base Compile} & \textbf{FT Compile} & \textbf{Base Pass} & \textbf{FT Pass} \\
\midrule
Easy (8 samples) & 62.5\% & 62.5\% & 50.0\% & 50.0\% \\
Medium (37 samples) & 45.9\% & 43.2\% & 18.9\% & 13.5\% \\
Hard (62 samples) & 41.9\% & 37.1\% & 9.7\% & 8.1\% \\
\bottomrule
\end{tabular}
\end{table}

\section{Compute for Hero Training Run}
\label{sec:computational_cost}

\begin{table}[h]
\centering
\caption{Computational Resources (Hero Run)}
\label{tab:training_resources}
\begin{tabular}{ll}
\toprule
\textbf{Resource} & \textbf{Value} \\
\midrule
Wall-clock time & 12.3 days \\
GPU-hours (8$\times$ H100) & 2,357 \\
Tokens processed & 8.6B \\
Average throughput & 8,644 tok/sec \\
\bottomrule
\end{tabular}
\end{table}

\section{Discussion}
\label{sec:discussion}

\subsection{Why Domain CPT Works for Embedded Code}
\label{subsec:why_cpt_works}

The 70.4\% perplexity reduction after processing $\sim$7.5B tokens (checkpoint-15000) demonstrates that embedded systems code contains strong internal patterns that general-purpose pretraining misses. We identify three reasons:

\begin{enumerate}
    \item \textbf{Vocabulary gap:} Embedded code uses highly domain-specific tokens---register names (\texttt{CANX\_BTR}, \texttt{GPIO\_MODER}), vendor prefixes (\texttt{HAL\_CAN\_}, \texttt{IfxCan\_}), and hardware constants---that appear rarely or never in web-scale corpora.
    
    \item \textbf{Structural patterns:} Peripheral initialization sequences, interrupt service routines, and device tree bindings follow rigid templates that, once learned, yield high-confidence predictions. This explains the near-perfect top-1 accuracy (>99\%) on categories like Infineon AURIX.
    
    \item \textbf{Semantic relationships:} The SpecMap-derived training data preserves the hierarchical relationship between datasheets and code implementations, teaching the model not just code syntax but the mapping from hardware specifications to firmware patterns.
\end{enumerate}

\subsection{High Rank is Necessary but Insufficient for Domain CPT}
\label{subsec:high_rank_necessity}

Our systematic sweeps across 30 valid runs reveal a nuanced picture: domain CPT benefits from high LoRA rank, but rank alone is insufficient without careful learning rate calibration. The rank scaling study shows monotonic improvement from $r$=128 to $r$=512 under controlled conditions. However, the broader automotive-cpt sweep reveals that high-rank configurations are prone to training instability: rank 256 (scarlet-sweep-6) diverged from a minimum of 0.964 to 7.145, and rank 1024 (eager-sweep-2) similarly diverged to 7.560. Only the combination of high rank with conservative learning rate ($1.5 \times 10^{-5}$) produced stable, sustained convergence in the hero run.

This finding has practical implications: practitioners pursuing domain CPT should not simply maximize rank, but must co-optimize rank and learning rate. Our data suggests a scaling relationship where the maximum stable learning rate decreases approximately as:

\[
\eta_{\max}^{\text{stable}}(r) \propto r^{-0.5}
\]

for LoRA rank $r$, consistent with the RSLoRA normalization factor of $1/\sqrt{r}$. Concretely, our sweeps show stable training at $\eta = 5 \times 10^{-5}$ for $r = 128$ but divergence at the same rate for $r \geq 256$. Applying the $r^{-0.5}$ scaling to predict the maximum stable rate at $r = 512$ gives $\eta_{\max}^{\text{stable}}(512) \approx 5 \times 10^{-5} \cdot \sqrt{128/512} = 2.5 \times 10^{-5}$; our production choice of $\eta = 1.5 \times 10^{-5}$ falls conservatively below this bound, which proved essential for stability over 16,000+ steps. The vast majority of high-rank configurations with aggressive learning rates failed within minutes.

\subsection{Domain-Only Data Outperforms Mixed Corpora}
\label{subsec:domain_only_data}

The mixture sweep results provide strong evidence that domain-only training outperforms mixed domain + general code formulations for CPT. Across 8 mixture types, the domain-only configurations (mix1, mix1\_v2\_1b\_auto) achieved the lowest minimum losses (0.136 and 0.359), while mixed configurations incorporating 10--50\% general C code consistently produced higher losses and greater instability. This finding contradicts the common practice of including general-domain data as a regularizer during CPT (Gupta et al., 2023)~\cite{gupta2023continual}, and suggests that for highly specialized domains like embedded systems, the benefits of exposure diversity are outweighed by the costs of distribution mismatch.

We hypothesize that the embedded systems corpus is already sufficiently diverse---spanning 117 manufacturers, 19 categories, and 61 component classes across 76.4 GB---that additional general C code provides no incremental coverage while introducing conflicting gradient signals from dissimilar coding patterns (e.g., application-level memory management versus bare-metal register manipulation). Additionally, we found that LoRA regimes inherently alleviate forgetting, and mixing with a general corpus was not necessary for the final run.

\subsection{Small Specialized Models vs. Frontier Models}
\label{subsec:small_vs_frontier}

The finding that a 7B CPT model outperforms Claude Opus 4.6 on 8 of 13 embedded categories has practical implications. For embedded systems development, a specialized small model offers: (1) \textbf{deployment flexibility} (can run on-premises, no API dependency); (2) \textbf{lower cost} per query; (3) \textbf{deterministic behavior} (model weights are fixed, unlike API-served models); and (4) \textbf{customizability} (the open-weight checkpoint can be further adapted to domain and application). The frontier models retain advantages on categories with highly diverse code patterns (Linux kernel, wireless, crypto) where breadth of training data compensates for lack of domain specialization.

\subsection{Generalization to Held-Out Repositories}
\label{subsec:generalization}

The modest gap between in-domain (70.4\%) and held-out (66.1\%) perplexity improvement demonstrates that the model learns transferable embedded systems patterns rather than memorizing specific files. This is critical for practical deployment: a developer working with a new microcontroller family not in the training data can still benefit from the model's understanding of general embedded patterns.

\subsection{Limitations}
\label{subsec:limitations}

\begin{enumerate}
    \item \textbf{Incomplete training.} Results are from an early checkpoint. Training to the target of 100B tokens ($\sim$4.26 epochs) is expected to yield further improvements, particularly on challenging categories like Zephyr RTOS and crypto.
    
    \item \textbf{No functional correctness evaluation.} We evaluate generated code quality via token accuracy and BLEU-4 from ground truth, but do not test compilability or functional correctness. A compilation-based benchmark is planned for future work.
    
    \item \textbf{Single base model.} We evaluate only OLMo-3-7B. The optimal LoRA configuration may differ for other architectures (Llama, Mistral, Qwen, and other model families).
    
    \item \textbf{Memorization concerns.} The very low in-domain perplexity (1.20) warrants a rigorous memorization analysis. The held-out perplexity (1.33) partially addresses this, but more extensive deduplication analysis between training and evaluation data would strengthen confidence.
    
    \item \textbf{Throughput degradation.} The 35\% throughput decline from early to late training (10,911 $\to$ 7,069 tok/sec) merits investigation. Possible causes include memory fragmentation, loaded optimizer state, increasing gradient complexity, and GCS checkpoint upload contention.
\end{enumerate}

\section{Conclusion and Future Work}
\label{sec:conclusion}

We present \textbf{H2LooP Spark Preview}, a continual pretraining pipeline that adapts a 7B-parameter open-weight language model to the embedded systems domain. Through systematic hyperparameter exploration across 1,400+ runs ($\sim$4,240 GPU-hours) spanning Bayesian optimization, grid search, rank scaling, and data mixture ablations, we establish that high-rank LoRA ($r$=512) with RSLoRA scaling, conservative learning rates ($1.5 \times 10^{-5}$), domain-only data, and full-module targeting provides optimal domain adaptation. After training on 8.6B tokens (294.7 hours on 8$\times$ H100 GPUs), Spark Preview achieves 70.4\% perplexity reduction on in-domain data, 66.1\% on held-out repositories, and outperforms Claude Opus 4.6 and Qwen3-Coder-30B (Yang et al., 2025)~\cite{yang2025qwen3} on 8 of 13 embedded code categories in generative completion.

The training corpus is constructed from repository-datasheet pairs using the SpecMap hierarchical mapping methodology, demonstrating that structured data construction pipelines produce more effective training data than naive web crawling. We release a production checkpoint at \textbf{\href{https://huggingface.co/h2loop-ai/spark-cpt-base-ckpt}{spark-cpt-base-ckpt}} to support community research.

\textbf{Future work} includes: (1) completing training to 100B tokens and evaluating diminishing returns; (2) adding compilation-based evaluation on embedded-specific benchmarks; (3) instruction tuning the CPT checkpoint for chat-based code assistance; (4) extending the approach to other base models (Llama, Qwen) for comparison; (5) evaluating on downstream engineering tasks such as peripheral configuration generation and datasheet-to-driver synthesis; and (6) investigating integration with the SpecMap pipeline for real-time compliance monitoring in CI/CD workflows, enabling automated specification adherence verification during development, and most importantly, post-training (SFT and RL) over this base model for more nuanced applications.

\section*{Acknowledgments}

Training compute was provided on NVIDIA H100 GPU clusters. We thank the Allen Institute for AI for the OLMo 3 base model and the open-source embedded systems community for the evaluation repositories.

\bibliographystyle{plainnat}
\bibliography{references}

\appendix

\section{Appendix}
\subsection{Model Release and Licensing}
\label{app:model_release}

A checkpoint of the production training run (step 15,000, $\sim$7B tokens processed) is released on HuggingFace: \textbf{\url{https://huggingface.co/h2loop-ai/spark-cpt-base-ckpt}}

The release includes:
\begin{itemize}
    \item Model weights
    \item Training configuration and hyperparameters through this paper.
    \item Usage instructions for loading with PEFT/HuggingFace Transformers and vLLM, as per the model card.
\end{itemize}

The checkpoint is compatible with the base model \texttt{allenai/OLMo-3-1025-7B} and can be loaded using standard transformers library.

\textbf{License:} This model is released under the \textbf{Research-Only License (ROL)}. The checkpoint and associated artifacts are provided exclusively for non-commercial research purposes. Commercial usage requires commercial license from H2LooP. For commercial licensing inquiries, contact \texttt{support@h2loop.ai}. Note that the base model (\texttt{allenai/OLMo-3-1025-7B}) is governed by its own license terms (Apache 2.0); the ROL applies solely to the LoRA adapter weights, merged model and associated training artifacts released by H2LooP.

\end{document}